\pdfoutput=1

\documentclass[11pt]{article}

\usepackage{acl}

\usepackage{times}
\usepackage{latexsym}

\usepackage[T1]{fontenc}

\usepackage[utf8]{inputenc}

\usepackage{microtype}

\usepackage{inconsolata}

\usepackage{url}
\usepackage{bbding}
\usepackage{pifont}
\definecolor{green}{HTML}{009B55}
\usepackage{microtype}
\usepackage{graphicx}
\usepackage{subfigure}
\usepackage{booktabs} 
\usepackage{xspace}
\usepackage[ruled,vlined]{algorithm2e}
\usepackage{macros}
\usepackage{wrapfig}
\usepackage{lipsum}
\usepackage{amsfonts}       
\usepackage{nicefrac}       
\usepackage{xcolor}         
\usepackage{hyperref}
\usepackage{multicol}
\usepackage{multirow}
\usepackage{listings}
\usepackage{pythonhighlight}
\usepackage{xcolor, colortbl}  
\definecolor{nblue}{cmyk}{0.95,0.0,0.2,0.2}
\newcommand{\blue}[1]{\textcolor{nblue}{\small #1}}
\usepackage{makecell}
\definecolor{yaleblue}{rgb}{0.06, 0.3, 0.57}
\definecolor{ruddybrown}{rgb}{0.73, 0.4, 0.16}

\newcommand{\method}{\texttt{EHRAgent}\xspace}

\usepackage{listings}
\usepackage{pythonhighlight}
\usepackage{fancyvrb}
\RecustomVerbatimCommand{\VerbatimInput}{VerbatimInput}{fontsize=\footnotesize,
 frame=single,  
 framesep=0.5em, 
 labelposition=topline,
}
\usepackage{tcolorbox}
\tcbuselibrary{listings}

\title{\method: Code Empowers Large Language Models for
Few-shot \\ Complex Tabular Reasoning on Electronic Health Records}

\author{Wenqi Shi$^1$\thanks{~~Equal contribution.}  \quad Ran Xu$^2$\footnotemark[1] \quad Yuchen Zhuang$^1$ \quad Yue Yu$^1$ \quad Jieyu Zhang$^3$ \\ \bf \quad Hang Wu$^1$ \quad Yuanda Zhu$^1$ \quad Joyce Ho$^2$ \quad Carl Yang$^2$ \quad May D. Wang$^1$ \\  
$^1$ Georgia Institute of Technology  \quad  $^2$ Emory University  \quad $^3$ University of Washington \\
\texttt{\{wqshi,yczhuang,yueyu,hangwu,yzhu94,maywang\}@gatech.edu}, \\ \texttt{\{ran.xu,joyce.c.ho,j.carlyang\}@emory.edu}, \ \texttt{jieyuz2@cs.washington.edu}
}

\begin{document}
\maketitle

\begin{abstract}
Clinicians often rely on data engineers to retrieve complex patient information from electronic health record (EHR) systems, a process that is both inefficient and time-consuming. 
We propose \method\footnote{Our implementation of \method is available at \url{https://github.com/wshi83/EhrAgent}.}, a large language model (LLM) agent empowered with accumulative domain knowledge and robust coding capability.
\method enables autonomous code generation and execution to facilitate clinicians in directly interacting with EHRs using natural language. 
Specifically, we formulate a multi-tabular reasoning task based on EHRs as a tool-use planning process, efficiently decomposing a complex task into a sequence of manageable actions with external toolsets. 
We first inject relevant medical information to enable \method to effectively reason about the given query, identifying and extracting the required records from the appropriate tables.
By integrating interactive coding and execution feedback, \method then effectively learns from error messages and iteratively improves its originally generated code. 
Experiments on three real-world EHR datasets show that \method outperforms the strongest baseline by up to 29.6\% in success rate, verifying its strong capacity to tackle complex clinical tasks with minimal demonstrations.

\end{abstract}

\section{Introduction}

\begin{figure}[t]
  \centering
  \includegraphics[width=0.99\linewidth]{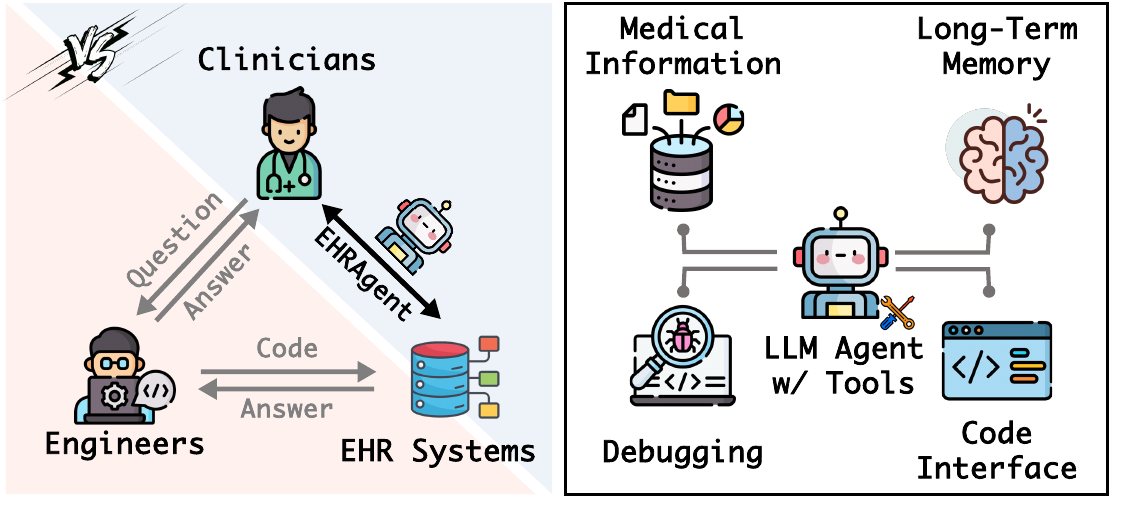}
  \caption{Simple and efficient interactions between clinicians and EHR systems with the assistance of LLM agents. Clinicians specify tasks in natural language, and the LLM agent autonomously generates and executes code to interact with EHRs (\textit{right}) for answers. It eliminates the need for specialized expertise or extra effort from data engineers, which is typically required when dealing with EHRs in existing clinical settings (\textit{left}).
}
  \vspace{-2ex}
  \label{fig:teaser}
\end{figure}

An electronic health record (EHR) is a digital version of a patient's medical history maintained by healthcare providers over time~\citep{gunter2005emergence}. 
In clinical research and practice, clinicians actively interact with EHR systems to access and retrieve patient data, ranging from detailed individual-level records to comprehensive population-level insights~\citep{cowie2017electronic}.
The reliance on pre-defined rule-based conversion systems in most EHRs often necessitates additional training or assistance from data engineers for clinicians to obtain information beyond these rules~\citep{mandel2016smart,bender2013hl7}, leading to inefficiencies and delays that may impact the quality and timeliness of patient care.

Alternatively, an autonomous agent could facilitate clinicians to communicate with EHRs in natural languages, translating clinical questions into machine-interpretable queries, planning a sequence of actions, and ultimately delivering the final responses. 
Compared to existing EHR management that relies heavily on human effort, the adoption of autonomous agents holds great potential to efficiently simplify workflows and reduce workloads for clinicians (Figure~\ref{fig:teaser}).
Although several supervised learning approaches~\citep{lee2022ehrsql,wang2020text} have been explored to automate the translation of clinical questions into corresponding machine queries, such systems require extensive training samples with fine-grained annotations, which are both expensive and challenging to obtain.

Large language models (LLMs)~\citep{openai2023gpt,palm2} bring us one step closer to autonomous agents with extensive knowledge and substantial instruction-following abilities from diverse corpora during pretraining.
LLM-based autonomous agents have demonstrated remarkable capabilities in problem-solving, such as reasoning~\citep{wei2022chain}, planning~\citep{yao2022react}, and memorizing~\citep{wang2023augmenting}. 
One particularly notable capability of LLM agents is tool-usage~\citep{schick2023toolformer,qin2023tool}, where they can utilize external tools (\eg, calculators, APIs, \etc), interact with environments, and generate action plans with intermediate reasoning steps that can be executed sequentially towards a valid solution~\citep{wu2023autogen,zhang2023ecoassistant}.

\begin{figure}[t]
	\centering
	\subfigure[\# Rows(k) per Table]{
		\includegraphics[width=0.46\linewidth]{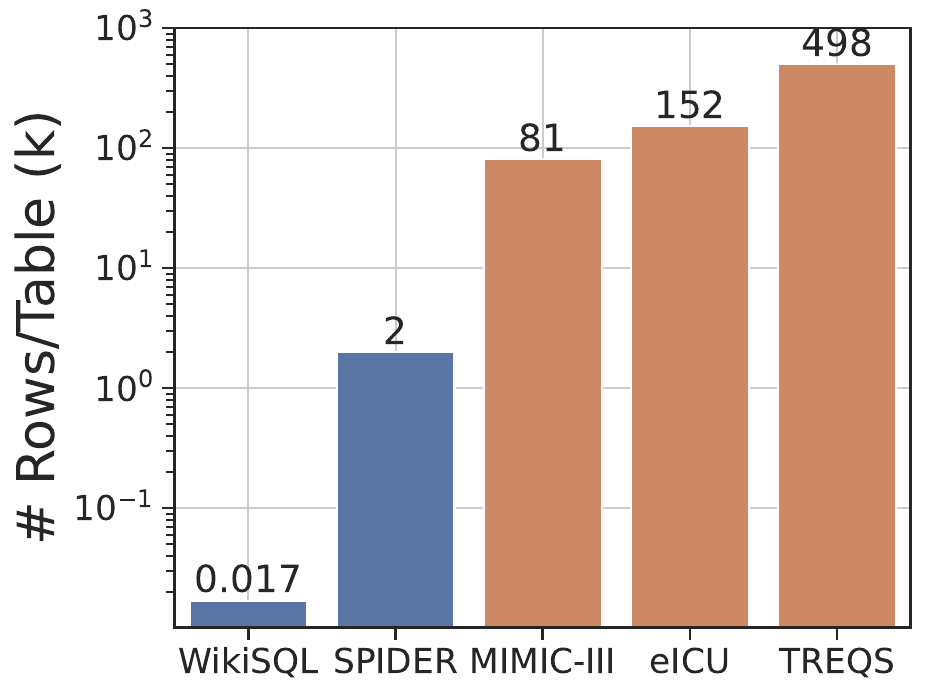}
		\label{fig:dataset-rows}
	} 
     \subfigure[\# Tables per Question]{
		\includegraphics[width=0.46\linewidth]{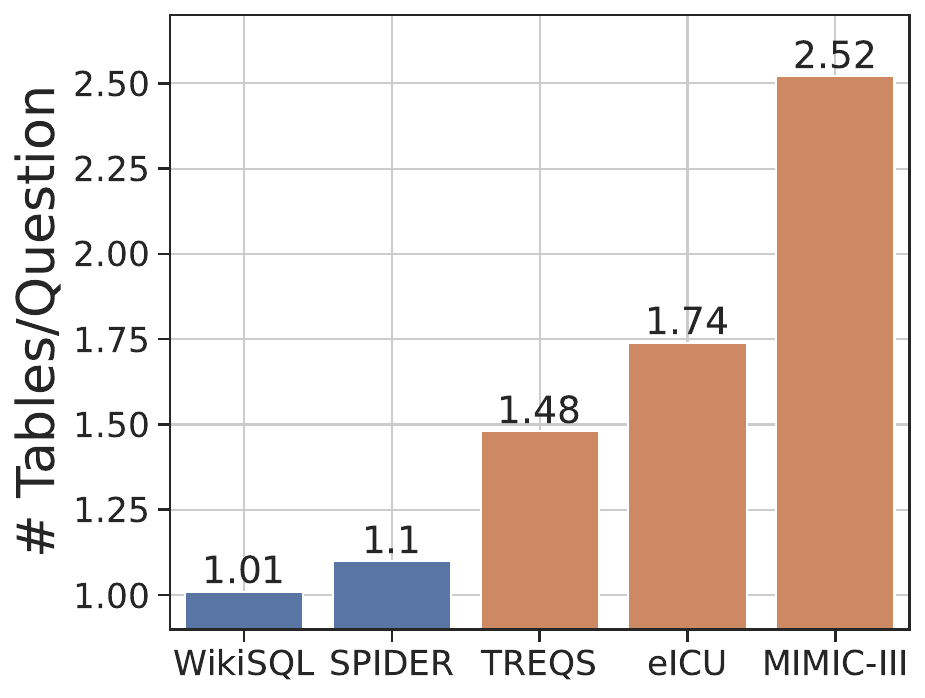}
		\label{fig:dataset-tables}
	}
      \vspace{-1ex}
	\caption{Compared to general domain tasks (\textcolor{yaleblue}{blue}) such as WikiSQL~\citep{zhong2017seq2sql} and SPIDER~\citep{yu-etal-2018-spider}, multi-tabular reasoning tasks within EHRs (\textcolor{ruddybrown}{orange}) typically involve a significantly larger number of records per table and necessitate querying multiple tables to answer each question, thereby requiring more advanced reasoning and problem-solving capabilities. 
}
 \vspace{-2ex}
\label{fig:compare-dataset}
\end{figure}

Despite their success in general domains, LLMs have encountered unique and significant challenges in the medical domain~\citep{jiang2023health,yang2022large,moor2023foundation}, especially when dealing with individual EHR queries that require advanced reasoning across a vast number of records within multiple tables~\citep{li2024can,lee2022ehrsql} (Figure~\ref{fig:compare-dataset}).
First, given the constraints in both the volume and specificity of training data within the medical field~\citep{thapa2023chatgpt}, LLMs still struggle to identify and extract relevant information from the appropriate tables and records within EHRs, due to insufficient knowledge and understanding of their complex structure and content.
Second, EHRs are typically large-scale relational databases containing vast amounts of tables with comprehensive administrative and clinical information (\eg, 26 tables of 46K patients in MIMIC-III).  
Moreover, 
real-world clinical tasks derived from individual patients or specific groups are highly diverse and complex, requiring multi-step or complicated operations.


To address these limitations, we propose \method, an autonomous LLM agent with external tools and code interface for improved multi-tabular reasoning across EHRs. 
We translate the EHR question-answering problem into a tool-use planning process -- generating, executing, debugging, and optimizing a sequence of code-based actions.
Firstly, to overcome the lack of domain knowledge in LLMs, we instruct \method to integrate query-specific medical information for effectively reasoning from the given query and locating the query-related tables or records.
Moreover, we incorporate long-term memory to continuously maintain a set of successful cases and dynamically select the most relevant few-shot examples, in order to effectively learn from and improve upon past experiences.
Secondly, we establish an interactive coding mechanism, which involves a multi-turn dialogue between the code planner and executor, iteratively refining the generated code-based plan for complex multi-hop reasoning.
Specifically, \method optimizes the execution plan by incorporating environment feedback and delving into error messages to enhance debugging proficiency.

We conduct extensive experiments on three large-scale real-world EHR datasets to validate the empirical effectiveness of \method, with a particular focus on challenging tasks that reflect diverse information needs and align with real-world application scenarios. 
In contrast to traditional supervised settings~\cite{lee2022ehrsql,wang2020text} that require over 10K training samples with manually crafted annotations, \method demonstrates its efficiency by necessitating only four demonstrations.
Our findings suggest that \method improves multi-tabular reasoning on EHRs through autonomous code generation and execution, leveraging accumulative domain knowledge and interactive environmental feedback.

Our main contributions are as follows:

$\bullet$ We propose \method, an LLM agent augmented with external tools and domain knowledge, to solve few-shot multi-tabular reasoning derived from EHRs with only four demonstrations; 

$\bullet$ Planning with a code interface, \method formulates a complex clinical problem-solving process as an executable code plan of action sequences, along with a code executor;

$\bullet$ We introduce interactive coding between the LLM agent and code executor, iteratively refining plan generation and optimizing code execution by examining environmental feedback in depth;

$\bullet$ Experiments on three EHR datasets show that \method improves the strongest baseline on multi-hop reasoning by up to 29.6\% in success rate.


\begin{figure*}[t]
  \centering
  \includegraphics[width=0.99\linewidth]{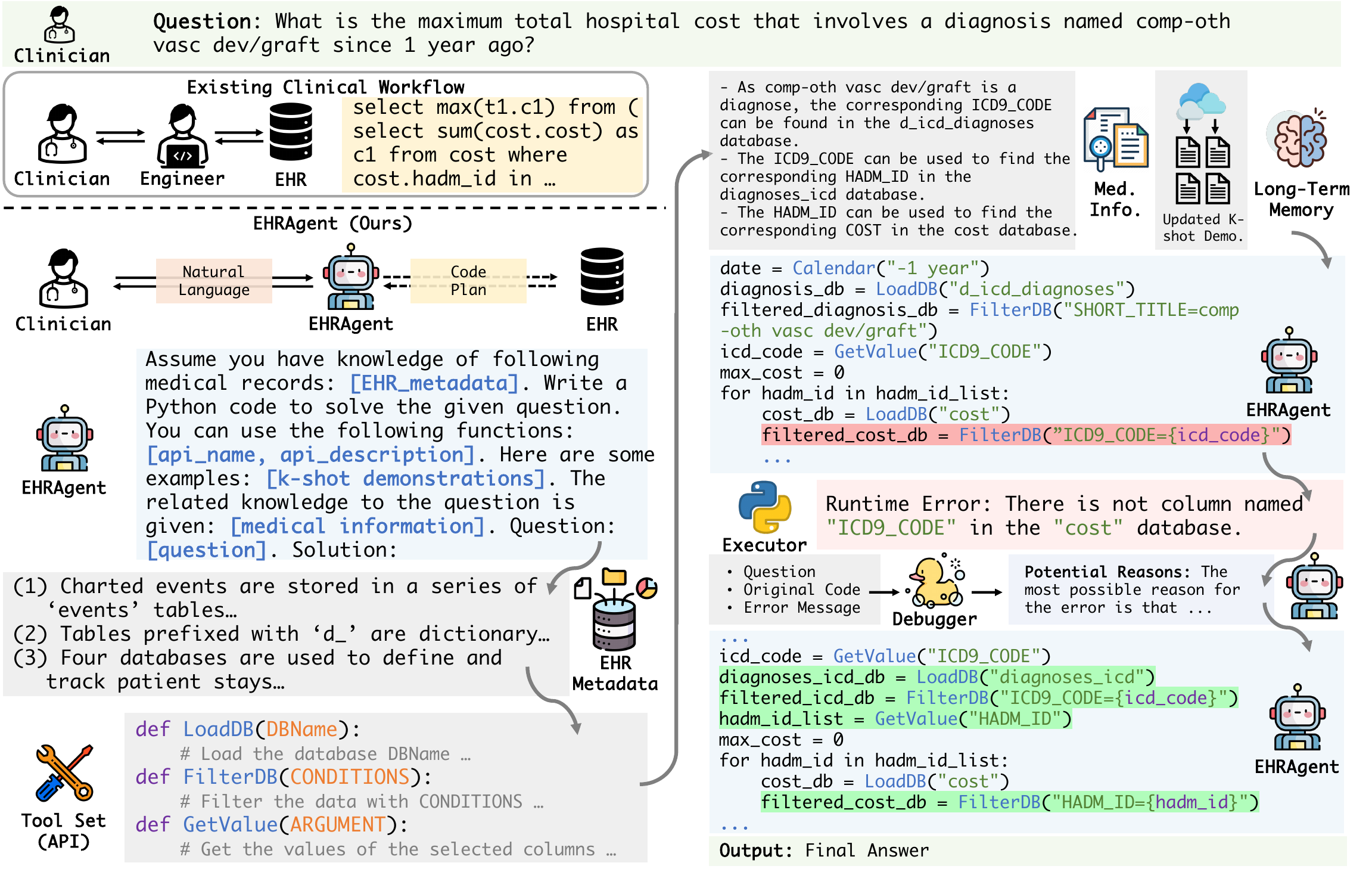}
  \caption{Overview of our proposed LLM agent, \method, for complex few-shot tabular reasoning tasks on EHRs. 
  Given an input clinical question based on EHRs, \method decomposes the task and generates a plan (\ie, code) based on (a) metadata (\ie, descriptions of tables and columns in EHRs), (b) tool function definitions, (c) few-shot examples, and (d) domain knowledge (\ie, integrated medical information). Upon execution, \method iteratively debugs the generated code following the execution errors and ultimately generates the final solution.
  }
  \vspace{-1ex}
  \label{fig:overview}
\end{figure*}

\section{Preliminaries}

\noindent\textbf{Problem Formulation.}
In this work, we focus on addressing health-related queries by leveraging information from structured EHRs. The reference EHR, denoted as $\mathcal{R}=\{R_0,R_1,\cdots\}$, comprises multiple tables, while $\mathcal{C}=\{C_0,C_1,\cdots\}$ corresponds to the column descriptions within $\mathcal{R}$. For each given query in natural language, denoted as $q$, our goal is to extract the final answer by utilizing the information within both $\mathcal{R}$ and $\mathcal{C}$.

\noindent\textbf{LLM Agent Setup.}
We further formulate the planning process for LLMs as autonomous agents in EHR question answering.
For initialization, the LLM agent is equipped with a set of pre-built tools $\mathcal{M}=\{M_0,M_1,\cdots\}$ to interact with and address queries derived from EHRs $\mathcal{R}$.
Given an input query $q\in\mathcal{Q}$ from the task space $\mathcal{Q}$, the objective of the LLM agent is to design a $T$-step execution plan $P=(a_1,a_2,\cdots,a_T)$, with each action $a_t$ selected from the tool set $a_t\in\mathcal{M}$.
Specifically, we generate the action sequences (\ie, plan) by prompting the LLM agent following a policy $p_q\sim \pi(a_1,\cdots,a_{T_q}|q;\mathcal{R},\mathcal{M}):
\mathcal{Q}\times\mathcal{R}\times\mathcal{M}\to\Delta(\mathcal{M})^{T_q}$, where $\Delta(\cdot)$ is a probability simplex function.
The final output is obtained by executing the entire plan $y\sim \rho(y|q,a_1,\cdots,a_{T_q})$, where $\rho$ is a plan executor interacting with EHRs.

\noindent\textbf{Planning with Code Interface.}
To mitigate ambiguities and misinterpretations in plan generation, an increasing number of LLM agents~\citep{gao2023pal, liang2023code, sun2023adaplanner,chen2022program,zhuang2024toolchain} employ code prompts as planner interface instead of natural language prompts.
The code interface enables LLM agents to formulate an executable code plan as action sequences, intuitively transforming natural language question-answering into iterative coding~\citep{yang2023intercode}.
Consequently, the planning policy $\pi(\cdot)$ turns into a code generation process, with a code execution as the executor $\rho(\cdot)$. 
We then track the outcome of each interaction back to the LLM agent, which can be either a successful execution result or an error message, to iteratively refine the generated code-based plan. 
This interactive process, a multi-turn dialogue between the planner and executor, takes advantage of the advanced reasoning capabilities of LLMs to optimize plan refinement and execution.


\section{\method: LLMs as Medical Agents}

In this section, we present \method (Figure~\ref{fig:overview}), an LLM agent that enables multi-turn interactive coding to address multi-hop reasoning tasks on EHRs.
\method comprises four key components:
(1) \textbf{Medical Information Integration:} We incorporate query-specific medical information for effective reasoning based on the given query, enabling \method to identify and retrieve the necessary tables and records for answering the question.
(2) \textbf{Demonstration Optimization through Long-Term Memory:} Using long-term memory, \method replaces original few-shot demonstrations with the most relevant successful cases retrieved from past experiences.
(3) \textbf{Interactive Coding with Execution Feedback:} \method harnesses LLMs as autonomous agents in a multi-turn conversation with a code executor.
(4) \textbf{Rubber Duck Debugging via Error Tracing:} Rather than simply sending back information from the code executor, \method thoroughly analyzes error messages to identify the underlying causes of errors through iterations until a final solution.
We summarize the workflow of \method in Algorithm~\ref{alg:method}.

\begin{algorithm}[t]
  \begin{small}
	\KwIn{$q$: input question; $\mathcal{R}$: reference EHRs; $\mathcal{C}_i$: column description of EHR $R_i$; $\mathcal{D}$: descriptions of EHRs $\mathcal{R}$; $T$: the maximum number of steps; $\mathcal{T}$: definitions of tool function; $\mathcal{L}$: long-term memory.} 
    Initialize $t\leftarrow 0$, $C^{(0)}(q)\leftarrow \varnothing$, $O^{(0)}(q)\leftarrow \varnothing$ \\
    \blue{// \textit{Medical Information Integration}}\\
    $\mathcal{I}=[\mathcal{D};\mathcal{C}_0;\mathcal{C}_1;\cdots]$\\
    $B(q)=\operatorname{LLM}([\mathcal{I};q])$ \\
    \blue{// \textit{Examples Retrieval from Long-Term Memory}}\\
    $\mathcal{E}(q)=\arg\operatorname{TopK}_{\max}(\operatorname{sim}(q,q_i|q_i\in\mathcal{L}))$ \\
    \blue{// \textit{Plan Generation}}\\
    $C^{(0)}(q)=\operatorname{LLM}([\mathcal{I};\mathcal{T};\mathcal{E}(q);q;B(q)])$\\
	\While{$t < T \And \operatorname{TERMINATE}\notin O^{(t)}(q)$}{
        \blue{// \textit{Code Execution}}\\
        $O^{(t)}(q)=\operatorname{EXECUTE}(C^{(t)}(q))$\\
        \blue{// \textit{Debugging and Plan Modification}}\\
        $C^{(t+1)}(q)=\operatorname{LLM}(\operatorname{DEBUG}(O^{(t)}(q)))$\\
        $t\leftarrow t+1$
      }
	\KwOut{Final answer (solved) or error message (unsolved) from $O^{(t)}(q)$.}
  \end{small}
  \caption{Overview of \method.}
  \label{alg:method}
\end{algorithm}

\subsection{Medical Information Integration}
\label{sec:mii}

Clinicians frequently pose complex inquiries that necessitate advanced reasoning across multiple tables and access to a vast number of records within a single query.
To accurately identify the required tables, we first incorporate query-specific medical information (\ie, domain knowledge) into \method to develop a comprehensive understanding of the query within a limited context length.
Given an EHR-based clinical question $q$ and the reference EHRs $\mathcal{R}=\{R_0,R_1,\cdots\}$, the objective of information integration is to generate the domain knowledge most relevant to $q$, thereby facilitating the identification and location of potential useful references within $\mathcal{R}$.
For example, given a query related to `\emph{Aspirin}', we expect LLMs to locate the drug `\emph{Aspirin}' at the PRESCRIPTION table, under the \textit{prescription\_name} column in the EHR.

To achieve this, we initially maintain a thorough metadata $\mathcal{I}$ of all the reference EHRs, including overall data descriptions $\mathcal{D}$ and the detailed column descriptions $\mathcal{C}_i$ for each individual EHR $R_i$, expressed as $\mathcal{I}=[\mathcal{D};\mathcal{C}_0;\mathcal{C}_1;\cdots]$.
To further extract additional background knowledge essential for addressing the complex query $q$, we then distill key information from the detailed introduction $\mathcal{I}$. Specifically, we directly prompt LLMs to generate the relevant information $B(q)$ based on demonstrations, denoted as $B(q)=\operatorname{LLM}([\mathcal{I};q])$. 


\subsection{Demonstration Optimization through Long-Term Memory}
\label{sec:ltm}
Due to the vast volume of information within EHRs and the complexity of the clinical questions, there exists a conflict between limited input context length and the number of few-shot examples.
Specifically, $K$-shot examples may not adequately cover the entire question types as well as the EHR information.
To address this, we maintain a long-term memory $\mathcal{L}$ for storing past successful code snippets and reorganizing few-shot examples by retrieving the most relevant samples from $\mathcal{L}$. 
Consequently, the LLM agent can learn from and apply patterns observed in past successes to current queries. The selection of $K$-shot demonstrations $\mathcal{E}(q)$ is defined as follows:
\begin{equation}
    \begin{aligned}
        \mathcal{E}(q)=\arg\operatorname{TopK}_{\max}(\operatorname{sim}(q,q_i|q_i\in\mathcal{L})),
    \end{aligned}
\end{equation}
where $\arg\operatorname{TopK}{\max}(\cdot)$ identifies the indices of the top $K$ elements with the highest values from $\mathcal{L}$, and $\operatorname{sim}(\cdot,\cdot)$ calculates the similarity between two questions, employing negative Levenshtein distance as the similarity metric.
Following this retrieval process, the newly acquired $K$-shot examples $\mathcal{E}(q)$ replace the originally predefined examples $\mathcal{E} = \{E_1, \cdots, E_K\}$. This updated set of examples serves to reformulate the prompt, guiding \method in optimal demonstration selection by leveraging accumulative domain knowledge.

\subsection{Interactive Coding with Execution}
We then introduce interactive coding between the LLM agent (\ie, code generator) and code executor to facilitate iterative plan refinement.
\method integrates LLMs with a code executor in a multi-turn conversation. The code executor runs the generated code and returns the results to the LLM. 
Within the conversation, \method navigates the subsequent phase of the dialogue, where the LLM agent is expected to either (1) continue to iteratively refine its original code in response to any errors encountered or (2) finally deliver a conclusive answer based on the successful execution outcomes.

\noindent \textbf{LLM Agent.} To generate accurate code snippets $C(q)$ as solution plans for the query $q$, we prompt the LLM agent with a combination of the EHR introduction $\mathcal{I}$, tool function definitions $\mathcal{T}$, a set of $K$-shot examples $\mathcal{E}(q)$ updated by long-term memory, the input query $q$, and the integrated medical information relevant to the query $B(q)$:
\begin{equation}\label{eq:c}
    \begin{aligned}
        C(q)=\operatorname{LLM}([\mathcal{I};\mathcal{T};\mathcal{E}(q);q;B(q)]).
    \end{aligned}
\end{equation}
We develop the LLM agent to (1) generate code within a designated coding block as required, (2) modify the code according to the outcomes of its execution, and (3) insert a specific code “TERMINATE” at the end of its response to indicate the conclusion of the conversation.

\noindent \textbf{Code Executor.} The code executor automatically extracts the code from the LLM agent's output and executes it within the local environment: $O(q)=\operatorname{EXECUTE}(C(q))$.
After execution, it sends back the execution results to the LLM agent for potential plan refinement and further processing.
Given the alignment of empirical observations and Python's inherent modularity with tool functions\footnote{We include additional analysis in Appendix~\ref{app:python} to further justify the selection of primary programming language.}, we select Python 3.9 as the primary coding language for interactions between the LLM agent and the code executor.

\subsection{Rubber Duck Debugging via Error Tracing}
Our empirical observations
indicate that LLM agents tend to make slight modifications to the code snippets based on the error message without further debugging. In contrast, human programmers often delve deeper, identifying bugs or underlying causes by analyzing the code implementation against the error descriptions~\citep{chen2023teaching}.
Inspired by this, we integrate a `rubber duck debugging' pipeline with error tracing to refine plans with the LLM agent. 
Specifically, we provide detailed trace feedback, including error type, message, and location, all parsed from the error information by the code executor.
Subsequently, this error context is presented to a `rubber duck' LLM, prompting it to generate the most probable causes of the error. The generated explanations are then fed back into the conversation flow, aiding in the debugging process.
For the $t$-th interaction between the LLM agent and the code executor, the process is as follows:
\begin{equation}
    \begin{aligned}
        & O^{(t)}(q)=\operatorname{EXECUTE}(C^{(t)}(q)),\\
        & C^{(t+1)}(q)=\operatorname{LLM}(\operatorname{DEBUG}(O^{(t)}(q))).
    \end{aligned}
\end{equation}
The interaction ends either when a `TERMINATE' signal appears in the generated messages or when $t$ reaches a pre-defined threshold of steps $T$.

\section{Experiments}
\label{sec:exp}

\begin{table*}[t]
\centering
\fontsize{8}{10}\selectfont\setlength{\tabcolsep}{0.3em}
\resizebox{\linewidth}{!}{
\begin{tabular}{@{}l|cccc>{\columncolor{pink!10}}c>{\columncolor{blue!5}}c|ccc>{\columncolor{pink!10}}c>{\columncolor{blue!5}}c|ccc>{\columncolor{pink!10}}c>{\columncolor{blue!5}}c@{}}
\toprule
\textbf{Dataset ($\rightarrow$) }  & \multicolumn{6}{c|}{\textbf{MIMIC-III}} & \multicolumn{5}{c|}{\textbf{eICU}} & \multicolumn{5}{c}{\textbf{TREQS}} \\ \midrule
\textbf{Complexity Level ($\rightarrow$)} & \textbf{I}     & \textbf{II}    & \textbf{III}   & \textbf{IV}    & \multicolumn{2}{c|}{\textbf{All}} & \textbf{I}    & \textbf{II}    & \textbf{III}     & \multicolumn{2}{c|}{\textbf{All}} & \textbf{I}    & \textbf{II}    & \textbf{III}     & \multicolumn{2}{c}{\textbf{All}}\\ 
\cmidrule{1-1}\cmidrule(lr){2-5} \cmidrule(lr){6-7} \cmidrule(lr){8-10} \cmidrule(lr){11-12} \cmidrule(lr){13-15} \cmidrule(lr){16-17}
\textbf{Methods ($\downarrow$) \slash Metrics ($\rightarrow$)} &  \multicolumn{4}{c}{\textbf{SR.}}  & \textbf{SR.} & \textbf{CR.} &  \multicolumn{3}{c}{\textbf{SR.}}  & \textbf{SR.} & \textbf{CR.} &  \multicolumn{3}{c}{\textbf{SR.}}  & \textbf{SR.} & \textbf{CR.}\\
\midrule
\multicolumn{17}{l}{\quad \emph{w/o Code Interface}}  \\\midrule 
CoT~\citep{wei2022chain}        & 29.33 & 12.88 & 3.08  & 2.11  & 9.58      & 38.23        & 26.73 & 33.00 & 8.33  & 27.34     & 65.65  & 11.22 & 9.15 & 0.00 & 9.84 & 54.02      \\
Self-Consistency~\cite{wang2022self} & 33.33 & 16.56 & 4.62 & 1.05 & 10.17 & 40.34 & 27.11 & 34.67 & 6.25 & 31.72 & 70.69 & 12.60 & 11.16 & 0.00 & 11.45 & 57.83 \\
Chameleon~\citep{lu2023chameleon}  & 38.67 & 14.11 & 4.62  & 4.21  & 12.77     & 42.76        & 31.09 & 34.68 & 16.67 & 35.06     & 83.41    & 13.58 & 12.72 & 4.55 & 12.25 & 60.34    \\
ReAct~\citep{yao2022react}      & 34.67 & 12.27 & 3.85  & 2.11  & 10.38     & 25.92        & 27.82 & 34.24 & 15.38 & 33.33     & 73.68  & 33.86 & 26.12 & 9.09 & 29.22 & 78.31     \\
Reflexion~\cite{shinn2023reflexion} & 41.05 & 19.31 & 12.57 & 11.96 & 19.48 & 57.07 & 38.08 & 33.33 & 15.38 & 36.72 & 80.00 & 35.04 & 29.91 & 9.09 &31.53 & 80.02\\\midrule
\multicolumn{17}{l}{\quad \emph{w/ Code Interface}}  \\\midrule 
LLM2SQL~\citep{nan2023evaluating}    & 23.68   & 10.64      &  6.98     &  4.83    &  13.10         &  44.83          &  20.48    &    25.13  &   12.50    &    23.28      & 51.72 & 39.61 & 36.43 & 12.73 & 37.89 & 79.22 \\
DIN-SQL~\citep{pourreza2024din}    & 49.51 & 44.22 & 36.25 & 21.85 & 38.45 &    81.72  &  23.49    &  26.13   & 12.50 & 25.00 & 55.00  & 41.34 & 36.38 & 12.73 & 38.05 & 82.73 \\
Self-Debugging~\cite{chen2023teaching} & 50.00 & 46.93 & 30.12 & 27.61 & 39.05 & 71.24 & 32.53 & 21.86 & \textbf{25.00} & 30.52 & 66.90 & 43.54 & 36.65 & 18.18 & 40.10 & 84.44 \\
AutoGen~\citep{wu2023autogen}    & 36.00 & 28.13 & 15.33 & 11.11 & 22.49     & 61.47        & 42.77 & 40.70 & 18.75 & 40.69     & 86.21   &46.65 & 19.42 & 0.00 & 33.13 & 85.38     \\
 \textbf{\method (Ours)}  & \textbf{71.58} & \textbf{66.34} & \textbf{49.70} & \textbf{49.14} & \textbf{58.97}     & \textbf{85.86}        & \textbf{54.82} & \textbf{53.52} & \textbf{25.00} & \textbf{53.10}      & \textbf{91.72}  & \textbf{78.94} & \textbf{61.16} & \textbf{27.27} & \textbf{69.70} & \textbf{88.02}     \\ \bottomrule
\end{tabular}
}
\caption{Main results of success rate (\ie, SR.) and completion rate (\ie, CR.) on MIMIC-III, eICU, and TREQS datasets. The complexity of questions increases from Level I (the simplest) to Level IV (the most difficult). }\label{tab:main}
\end{table*}

\subsection{Experiment Setup}
\noindent \textbf{Tasks and Datasets.} 
We evaluate \method on three publicly available structured EHR datasets, MIMIC-III~\citep{johnson2016mimic}, eICU~\citep{pollard_eicu_2018}, and TREQS~\citep{wang2020text} for multi-hop question and answering on EHRs. 
These questions originate from real-world clinical needs and cover a wide range of tabular queries commonly posed within EHRs. 
Our final dataset includes an average of 10.7 tables and 718.7 examples per dataset, with an average of 1.91 tables required to answer each question. 
We include additional dataset details in Appendix~\ref{app:dataset}.

\noindent \textbf{Tool Sets.}
To enable LLMs in complex operations such as calculations and information retrieval, we integrate external tools in \method during the interaction with EHRs.
Our toolkit can be easily expanded with natural language tool function definitions in a plug-and-play manner.
Toolset details are available in Appendix~\ref{app:toolset}.

\noindent \textbf{Baselines.}
We compare \method with nine LLM-based planning, tool use, and coding methods, including five baselines with natural language interfaces and four with coding interfaces. 
For a fair comparison, all baselines, including \method, utilize the same (a) EHR metadata, (b) tool definitions, and (c) initial few-shot demonstrations in the prompts by default.
We summarize their implementations in Appendix~\ref{app:baseline_attr}.

\noindent \textbf{Evaluation Protocol.} 
Following~\citet{yao2022react,sun2023adaplanner,shinn2023reflexion}, our primary evaluation metric is \emph{success rate}, quantifying the percentage of queries the model handles successfully.
Following \citet{xu2023tool,kirk2024improving}, we further assess \emph{completion rate}, which represents the percentage of queries that the model can generate executable plans (even not yield correct results).
We categorize input queries into complexity levels (I-IV) based on the number of tables involved in solution generation. 
We include more details in Appendix~\ref{app:complexity-level}.

\noindent \textbf{Implementation Details.}
We employ \texttt{GPT-4}~\citep{openai2023gpt} (version \texttt{gpt-4-0613}) as the base LLM model for all experiments. 
We set the temperature to 0 when making API calls to \texttt{GPT-4} to eliminate randomness and set the pre-defined threshold of steps ($T$) to 10.
Due to the maximum length limitations of input context in baselines (\eg, ReAct and Chameleon), we use the same initial four-shot demonstrations ($K=4$) for all baselines and \method to ensure a fair comparison.
Appendix~\ref{app:implementation} provides additional implementation details with prompt templates.

\subsection{Main Results}
Table~\ref{tab:main} summarizes the experimental results of \method and baselines on multi-tabular reasoning within EHRs. From the results, we have the following observations:

\noindent (1) \method significantly outperforms all the baselines on all three datasets with a performance gain of 19.92\%, 12.41\%, and 29.60\%, respectively. This indicates the efficacy of our key designs, namely interactive coding with environment feedback and domain knowledge injection, as they gradually refine the generated code and provide sufficient background information during the planning process. Experimental results with additional base LLMs are available in Appendix~\ref{app:gpt-35}.

\noindent (2) \emph{CoT}, \emph{Self-Consistency}, and \emph{Chameleon}
all neglect environmental feedback and cannot adaptively refine their planning processes. Such deficiencies hinder their performance in EHR question-answering scenarios, as the success rates for these methods on three datasets are all below 40\%.

\noindent (3) \emph{ReAct} and \emph{Reflexion} both consider environment feedback but are restricted to tool-generated error messages. Thus, they potentially overlook the overall planning process. 
Moreover, they both lack a code interface, which prevents them from efficient action planning, and results in lengthy context execution and lower completion rates.



\noindent (4) \emph{LLM2SQL} and \emph{DIN-SQL} leverage LLM to directly generate SQL queries for EHR question-answering tasks. However, the gain is rather limited, as the LLM still struggles to generate high-quality SQL codes for execution. Besides, the absence of the debugging module further impedes its overall performance on this challenging task.

\noindent (5) \emph{Self-Debugging} and \emph{AutoGen} present a notable performance gain over other baselines, as they leverage code interfaces and consider the errors from the coding environment, leading to a large improvement in the completion rate. 
However, as they fail to model medical knowledge or identify underlying causes from error patterns, their success rates are still sub-optimal.

\subsection{Ablation Studies}
\begin{table}[t]
\centering
\fontsize{8}{10}\selectfont\setlength{\tabcolsep}{0.3em}
\resizebox{\linewidth}{!}{
\begin{tabular}{@{}lcccc>{\columncolor{pink!10}}c>{\columncolor{blue!5}}c@{}}
\toprule
     \textbf{Complexity Level ($\rightarrow$)}     & \textbf{I}     & \textbf{II}    & \textbf{III}   & \textbf{IV}   & \multicolumn{2}{c}{\textbf{All}}    \\
    \cmidrule(lr){2-5} \cmidrule(lr){6-7}
    \textbf{Methods ($\downarrow$) \slash Metrics ($\rightarrow$)} &  \multicolumn{4}{c}{\textbf{SR.}}  & \textbf{SR.} & \textbf{CR.} \\
    \midrule
\textbf{\method}                              & \textbf{71.58} & \textbf{66.34} & \textbf{49.70}  & \textbf{49.14} & \textbf{58.97} & \textbf{85.86} \\ 
\quad w/o medical information       & 68.42 & 33.33 & 29.63 & 20.00    & 33.66 & 69.22 \\
\quad w/o long-term memory                  & 65.96 & 54.46 & 37.13 & 42.74 & 51.73 & 83.42 \\
\quad w/o interactive coding & 45.33 & 23.90  & 20.97 & 13.33 & 24.55 & 62.14 \\
\quad w/o rubber duck debugging  & 55.00    & 38.46 & 41.67 & 35.71 & 42.86 & 77.19 \\ \bottomrule
\end{tabular}
}
\caption{Ablation studies on success rate (\ie, SR.) and completion rate (\ie, CR.) under different question complexity (I-IV) on MIMIC-III dataset.}\label{tab:ablation}
\end{table}

Our ablation studies on MIMIC-III (Table~\ref{tab:ablation}) demonstrate the effectiveness of all four components in \method.
Interactive coding\footnote{For \method w/o interactive coding, we deteriorate from generating code-based to natural language-based plans and enable debugging based on error messages from tool execution.} is the most significant contributor across all complexity levels, which highlights the importance of code generation in planning and environmental interaction for refinement.
In addition, more challenging tasks benefits more from knowledge integration, indicating that comprehensive understanding of EHRs facilitates the complex multi-tabular reasoning in effective schema linking and reference (\eg, tables, columns, and condition values) identification.
Detailed analysis with additional settings and results is available in Appendix~\ref{app:abl}.

\subsection{Quantitative Analysis}
\begin{figure}[t]
	\centering
	\subfigure[success rate]{
		\includegraphics[width=0.46\linewidth]{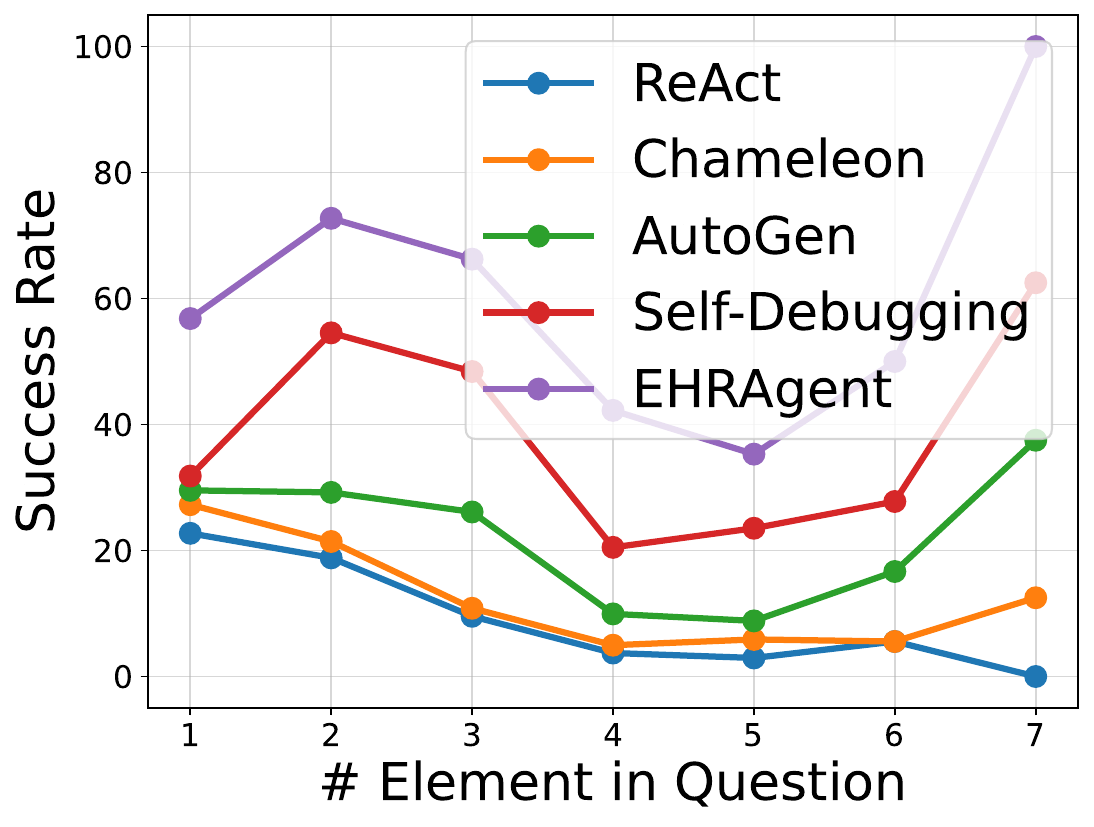}
		\label{fig:sr-qtag-m1}
	} 
     \subfigure[completion rate]{
		\includegraphics[width=0.46\linewidth]{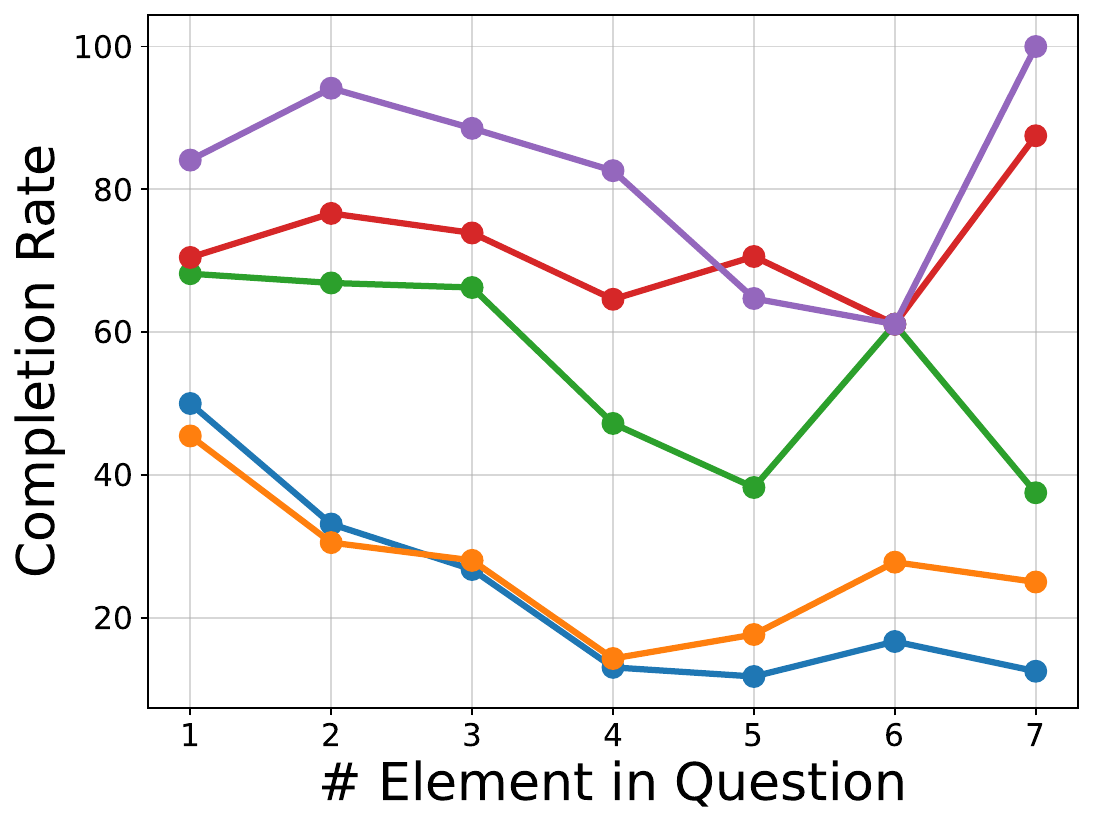}
		\label{fig:cr-qtag-m1}
	}
 
    \subfigure[success rate]{
		\includegraphics[width=0.46\linewidth]{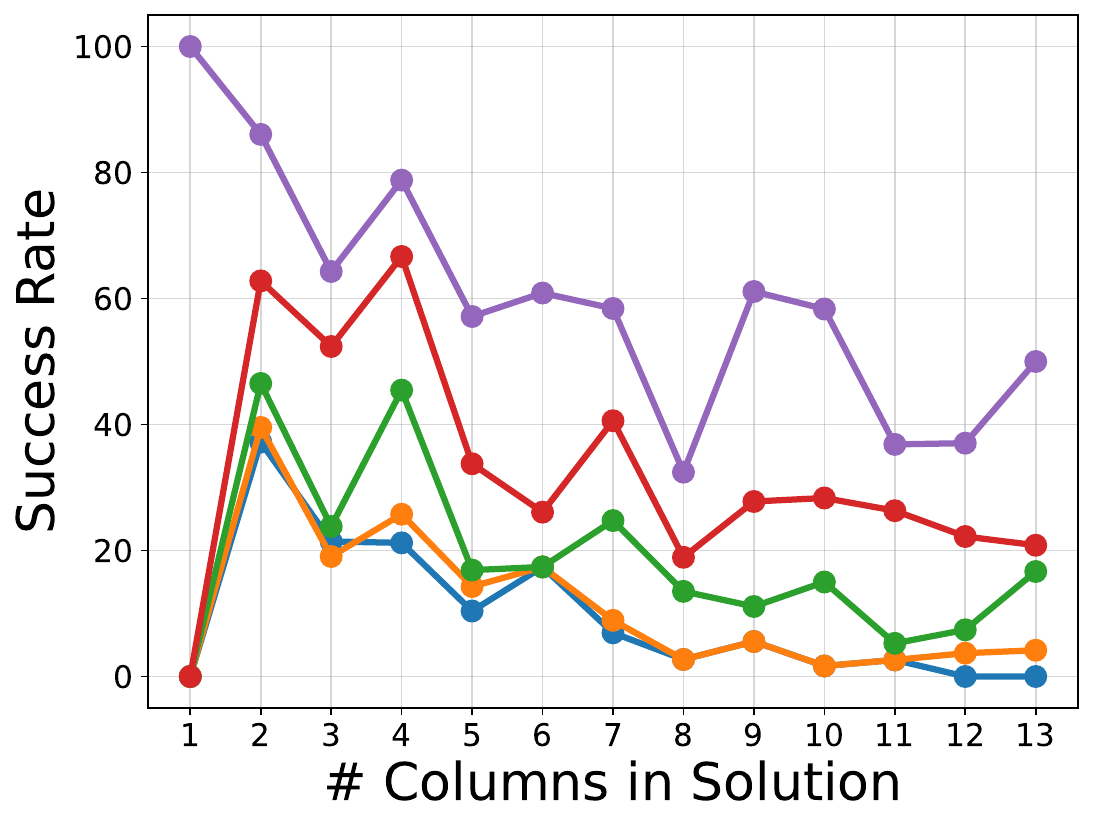}
		\label{fig:sr-qtag-m2}
	} 
     \subfigure[completion rate]{
		\includegraphics[width=0.46\linewidth]{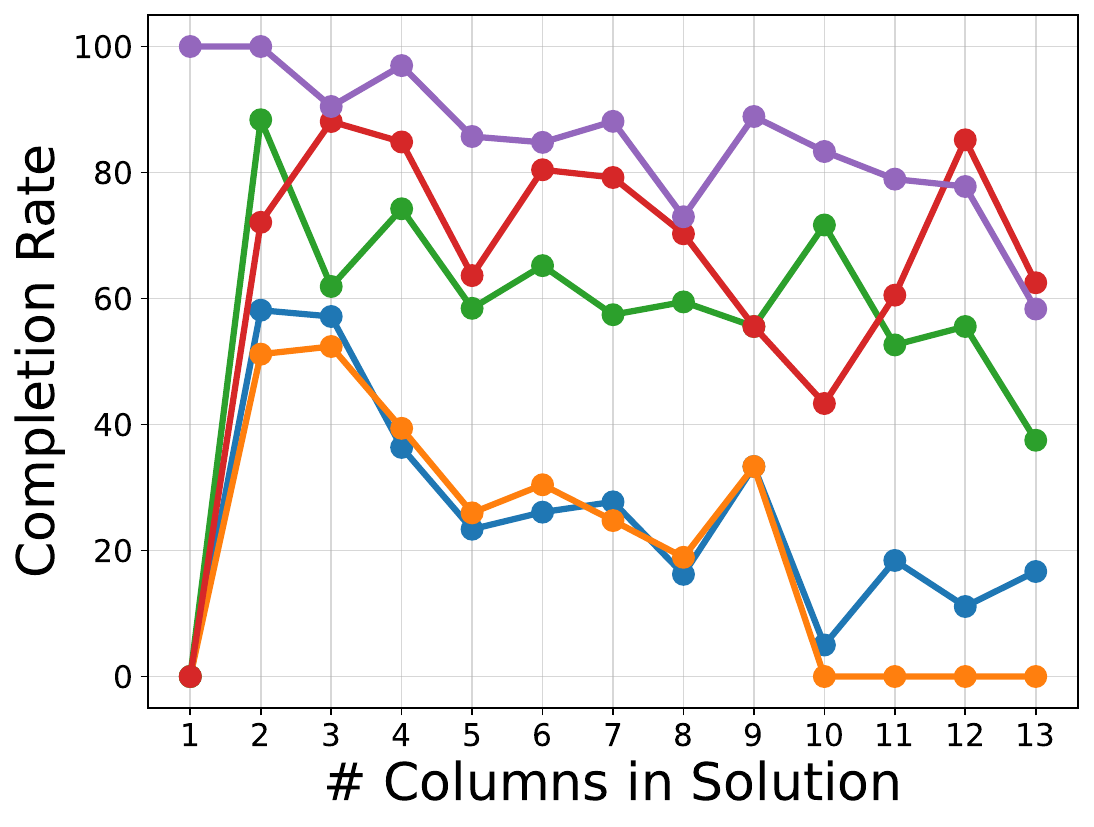}
		\label{fig:cr-qtag-m2}
	}
    \caption{Success rate and completion rate under different question complexity, measured by the number of elements (\ie, slots) in each question (\textit{upper}) and the number of columns involved in each solution (\textit{bottom}).}
\label{fig:qc}
\end{figure}

\noindent \textbf{Effect of Question Complexity.}
We take a closer look at the model performance by considering multi-dimensional measurements of question complexity, exhibited in Figure~\ref{fig:qc}. 
Although the performances of both \method and the baselines generally decrease with an increase in task complexity (either quantified as more elements in queries or more columns in solutions), \method consistently outperforms all the baselines at various levels of difficulty.
Appendix~\ref{app:qc-ana} includes additional analysis on the effect of various question complexities.

\begin{figure}[t]
	\centering
	\subfigure[success rate]{
		\includegraphics[width=0.46\linewidth]{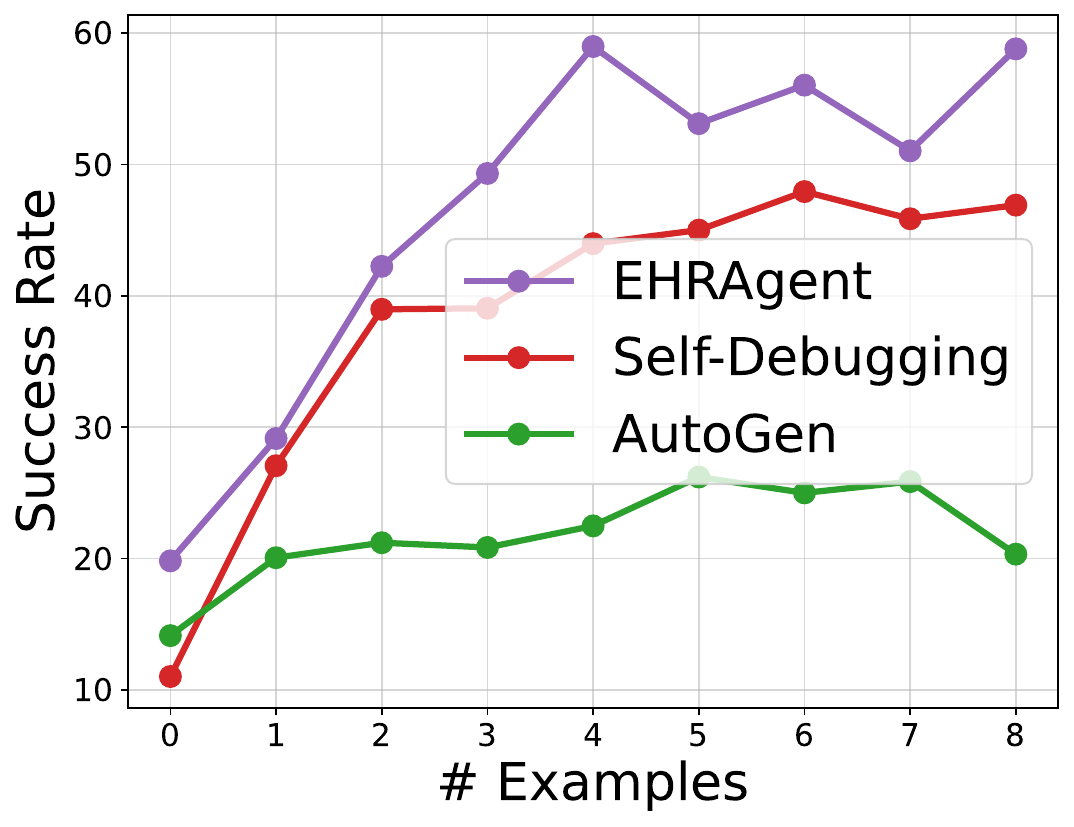}
		\label{fig:sr-se}
	} 
     \subfigure[completion rate]{
		\includegraphics[width=0.46\linewidth]{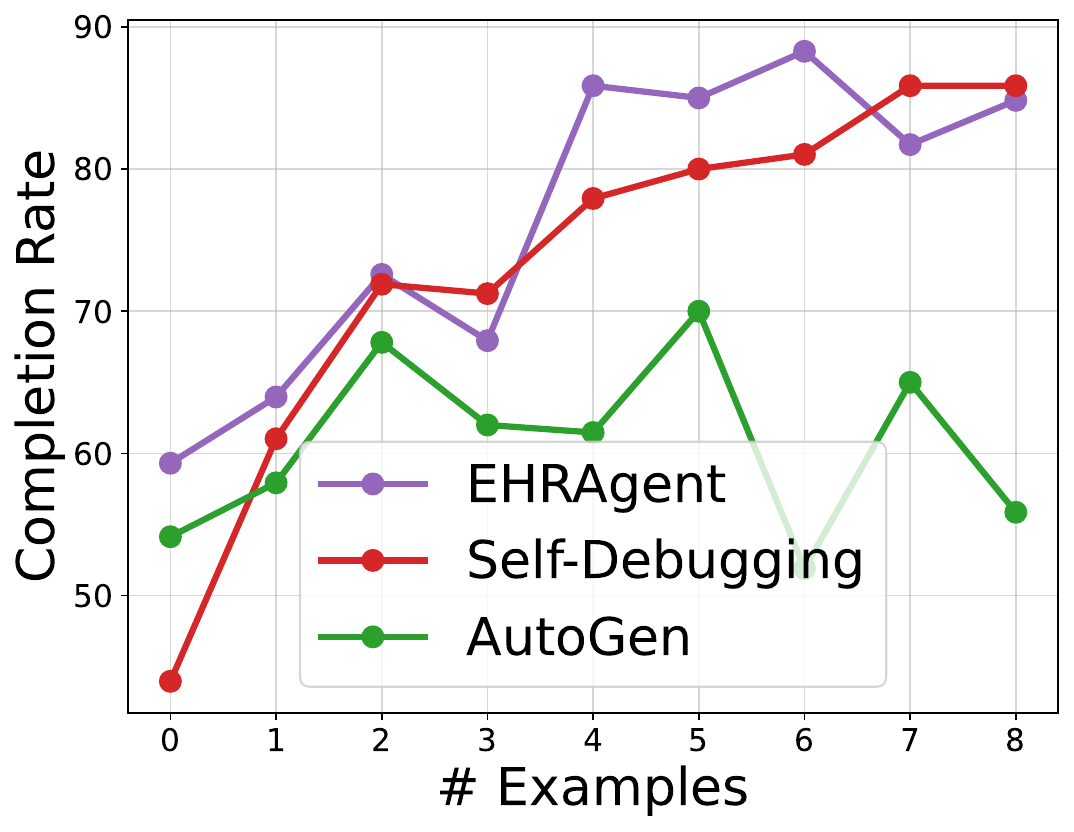}
		\label{fig:cr-se}
	}
	\caption{Success rate and completion rate under different numbers of demonstrations.}
\label{fig:se}
\end{figure}

\begin{figure}[t]
  \centering
  \includegraphics[width=0.99\linewidth]{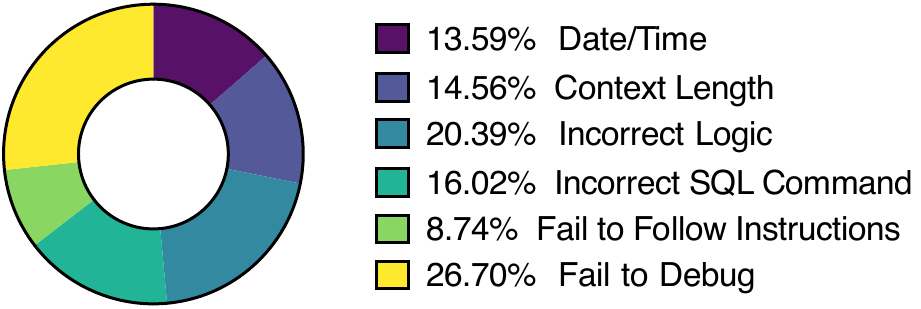}
  \caption{Percentage of mistake examples in different categories on MIMIC-III dataset.
}
  \label{fig:error}
\end{figure}

\noindent \textbf{Sample Efficiency.}
Figure~\ref{fig:se} illustrates the model performance \wrt $\ $number of demonstrations for \method and the two strongest baselines, AutoGen and Self-Debugging.
Compared to supervised learning like text-to-SQL~\citep{wang2020text,raghavan2021emrkbqa,lee2022ehrsql} that requires extensive training on over 10K samples with detailed annotations (\eg, manually generated corresponding code for each query), LLM agents enable complex tabular reasoning using a few demonstrations only.
One interesting finding is that as the number of examples increases, both the success and completion rate of AutoGen tend to decrease, mainly due to the context limitation of LLMs.
Notably, the performance of \method remains stable with more demonstrations, which may benefit from its integration of a `rubber duck' debugging module and the adaptive mechanism for selecting the most relevant demonstrations. 

\subsection{Error Analysis}
Figure~\ref{fig:error} presents a summary of error types identified in the solution generation process of \method based on the MIMIC-III, as determined through manual examinations and analysis. 
The majority of errors occur because the LLM agent consistently fails to identify the underlying cause of these errors within $T$-step trails, resulting in plans that are either incomplete or inexcusable.
Additional analysis of each error type is available in Appendix \ref{app:err}.

\subsection{Case Study}
Figure~\ref{fig:cs-main1} presents a case study of \method in interactive coding with environment feedback. 
The initial solution from LLM is unsatisfactory with multiple errors. 
Fortunately, \method is capable of identifying the underlying causes of errors by analyzing error messages and resolves multiple errors one by one through iterations.
We have additional case studies in Appendix~\ref{app:case}.

\begin{figure}[t]
  \centering
  \includegraphics[width=0.99\linewidth]{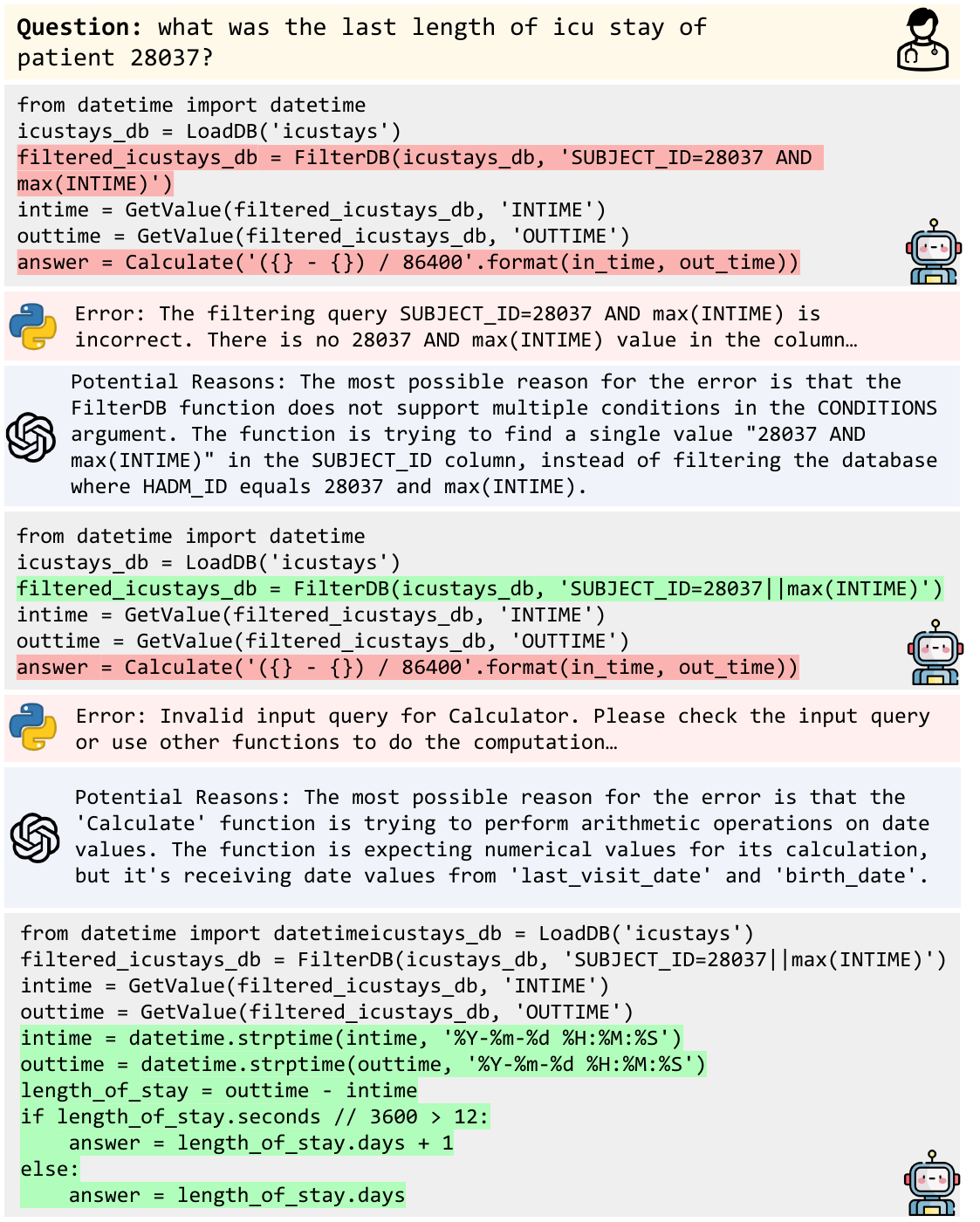}
  \vspace{-2ex}
  \caption{Case study of \method harnessing LLMs in a multi-turn conversation with a code executor, debugging with execution errors through iterations. 
  }
  \label{fig:cs-main1}
\end{figure}



\section{Related Work}
\noindent\textbf{Augmenting LLMs with External Tools.}
LLMs have rapidly evolved from text generators into core computational engines of autonomous agents, with advanced planning and tool-use capabilities~\citep{schick2023toolformer,shen2023hugginggpt,wang2023mint,yuan2023craft,yuan2024easytool,zhuang2023toolqa}.
LLM agents equip LLMs with planning capabilities~\citep{yao2023tree,gong2023mindagent} to decompose a large and hard task into multiple smaller and simpler steps for efficiently navigating complex real-world scenarios. 
By integrating with external tools, LLM agents access external APIs for additional knowledge beyond training data~\citep{lu2023chameleon,patil2023gorilla,qin2023toolllm,li-etal-2023-api,li2023camel}.
The disconnection between plan generation and execution, however, prevents LLM agents from effectively and efficiently mitigating error propagation and learning from environmental feedback~\citep{qiao2023making,shinn2023reflexion,yang2023intercode}. 
To this end, we leverage interactive coding to learn from dynamic interactions between the planner and executor, iteratively refining generated code by incorporating insights from error messages.
Furthermore, \method extends beyond the limitation of short-term memory obtained from in-context learning, leveraging long-term memory~\citep{sun2023adaplanner,zhang2023ecoassistant} by rapid retrieval of highly relevant and successful experiences accumulated over time.

\noindent\textbf{LLM Agents for Scientific Discovery.}
Augmenting LLMs with domain-specific tools, LLM agents have demonstrated capabilities of autonomous design, planning, and execution in accelerating scientific discovery~\citep{wang2023survey,wang2023scibench,wang2024executable,xi2023rise,zhao2023expel,cheung-etal-2024-polyie,gao2024empowering}, including organic synthesis~\citep{bran2023chemcrow}, material design~\citep{boiko2023autonomous}, and gene prioritization~\citep{jin2023genegpt}.
In the medical field, MedAgents~\citep{tang2023medagents}, a multi-agent collaboration framework, leverages role-playing LLM-based agents in a task-oriented multi-round discussion for multi-choice questions in medical entrance examinations.
Similarly, \citet{abbasian2023conversational} develop a conversational agent to enhance LLMs using external tools for general medical question-answering tasks.
Different from existing LLM agents in the medical domains that focus on improving tasks like multiple-choice question-answering, \method integrates LLMs with an interactive code interface, exploring complex few-shot tabular reasoning tasks derived from real-world EHRs through autonomous code generation and execution.

\vspace{-0.5ex}
\section{Conclusion}
\vspace{-0.5ex}
In this study, we develop \method, an LLM agent with external tools for few-shot multi-tabular reasoning on real-world EHRs.
Empowered by the emergent few-shot learning capabilities of LLMs, \method leverages autonomous code generation and execution for direct communication between clinicians and EHR systems.
We also improve \method by interactive coding with execution feedback, along with accumulative medical knowledge, thereby effectively facilitating plan optimization for multi-step problem-solving.
Our experiments demonstrate the advantages of \method over baseline LLM agents in autonomous coding and improved medical reasoning.

\section*{Limitation and Future Work}
\method holds considerable potential for positive social impact in a wide range of clinical tasks and applications, including but not limited to patient cohort definition, clinical trial recruitment, case review selection, and treatment decision-making support. Despite the significant improvement in model performance, we have identified several potential limitations of \method as follows:
\paragraph{Additional Execution Calls.}
We acknowledge that when compared to open-loop systems such as CoT, Self-Consistency, Chameleon, and LLM2SQL, which generate a complete problem-solving plan at the beginning without any adaptation during execution; \method, as well as other baselines that rely on environmental feedback like ReAct, Reflexion, Self-Debugging, and AutoGen, require additional LLM calls due to the multi-round conversation. However, such open-loop systems all overlook environmental feedback and cannot adaptively refine their planning processes. These shortcomings largely hinder their performance for the challenging EHR question-answering task, as the success rates for these methods on all three EHR datasets are all below 40\%. We can clearly observe the trade-off between performance and execution times. 
Although environmental feedback enhances performance, future work will focus on cost-effective improvements to balance performance and cost~\citep{zhang2023ecoassistant}.

\paragraph{Translational Clinical Research Considerations.}
Given the demands for privacy, safety, and ethical considerations in real-world clinical research and practice settings, our goal is to further advance \method by mitigating biases and addressing ethical implications, thereby contributing to the development of responsible artificial intelligence for healthcare and medicine. Furthermore, the adaptation and generalization of \method in low-resource languages is constrained by the availability of relevant resources and training data.
Due to limited access to LLMs' API services and constraints related to budget and computation resources, our current experiments are restricted to utilizing the Microsoft Azure OpenAI API service with the \texttt{gpt-3.5-turbo (0613)} and \texttt{gpt-4 (0613)} models.
As part of our important future directions, we plan to enhance \method by incorporating fine-tuned white-box LLMs, such as LLaMA-2~\citep{touvron2023llama}.

\paragraph{Completion Rate under Clinical Scenarios.}
Besides success rate (SR) as our main evaluation metric, we follow \citet{xu2023tool,kirk2024improving} and employ completion rate (CR) to denote the percentage of queries for which the model can generate executable plans, irrespective of whether the results are accurate.
However, it is important to note that a higher CR may not necessarily imply a superior outcome, especially in clinical settings. In such cases, it is generally preferable to acknowledge failure rather than generate an incorrect answer, as this could lead to an inaccurate diagnosis.
We will explore stricter evaluation metrics to assess the cases of misinformation that could pose a risk within clinical settings in our future work.

\section*{Privacy and Ethical Statement}
In compliance with the PhysioNet Credentialed Health Data Use Agreement 1.5.0\footnote{\url{https://physionet.org/about/licenses/physionet-credentialed-health-data-license-150/}}, we strictly prohibit the transfer of confidential patient data (MIMIC-III and eICU) to third parties, including through online services like APIs. To ensure responsible usage of Azure OpenAI Service based on the guideline\footnote{\url{https://physionet.org/news/post/gpt-responsible-use}}, we have opted out of the human review process by requesting the Azure OpenAI Additional Use Case Form\footnote{\url{https://aka.ms/oai/additionalusecase}}, which prevents third-parties (\eg, Microsoft) from accessing and processing sensitive patient information for any purpose. We continuously and carefully monitor our compliance with these guidelines and the relevant privacy laws to uphold the ethical use of data in our research and operations.


\section*{Acknowledgments}
We thank the anonymous reviewers and area chairs for their valuable feedback. This research was partially supported by Accelerate Foundation Models Academic Research Initiative from Microsoft Research.
This research was also partially supported by the National Science Foundation under Award Number 2319449 and Award Number 2312502, the National Institute Of Diabetes And Digestive And Kidney Diseases of the National Institutes of Health under Award Number K25DK135913, the Emory Global Diabetes Center of the Woodruff Sciences Center, Emory University.

\bibliography{anthology, custom}

\appendix
\section{Dataset and Task Details}
\label{app:dataset}

\subsection{Task Details}
We evaluate \method on three publicly available EHR datasets from two text-to-SQL medical question answering (QA) benchmarks~\citep{lee2022ehrsql}, EHRSQL\footnote{\url{https://github.com/glee4810/EHRSQL}} and TREQS\footnote{\url{https://github.com/wangpinggl/TREQS}}, built upon structured EHRs from MIMIC-III and eICU. 
EHRSQL and TREQS serve as text-to-SQL benchmarks for assessing the performance of medical QA models, specifically focusing on generating SQL queries for addressing a wide range of real-world questions gathered from over 200 hospital staff.
Questions within EHRSQL and TREQS, ranging from simple data retrieval to complex operations such as calculations, reflect the \textit{diverse} and \textit{complex} clinical tasks encountered by front-line healthcare professionals.
Dataset statistics are available in Table~\ref{tab:data}. 

\begin{table}[ht]
\centering
\fontsize{8}{10}\selectfont\setlength{\tabcolsep}{0.5em}
\begin{tabular}{@{}cccccc@{}}
\toprule
Dataset        & \# Examples & \# Table     & \# Row/Table         & \# Table/Q      \\ \midrule
MIMIC-III      & 580   & 17          &     81k          &   2.52                         \\
eICU           & 580      & 10          &    152k           &    1.74                        \\ 
TREQS          & 996     & 5          & 498k &   1.48     \\
\rowcolor{gray!16}\textbf{Average} & \textbf{718.7}  & \textbf{10.7} & \textbf{243.7k} & \textbf{1.91}  \\ \bottomrule
\end{tabular}
\caption{Dataset statistics.}
\label{tab:data}
\end{table}

\subsection{Question Complexity Level}
\label{app:complexity-level}
We categorize input queries into various complexity levels (levels I-IV for MIMIC-III and levels I-III for eICU and TREQS) based on the number of tables involved in solution generation.
For example, given the question `How many patients were given temporary tracheostomy?', the complexity level is categorized as II, indicating that we need to extract information from two tables (admission and procedure) to generate the solution.
Furthermore, we also conduct a performance analysis (see Figure~\ref{fig:qc}) based on additional evaluation metrics related to question complexity, including (1) the number of elements (\ie, slots) in each question and (2) the number of columns involved in each solution.
Specifically, elements refer to the slots within each template that can be populated with pre-defined values or database records. 

\subsection{MIMIC-III}
MIMIC-III~\citep{johnson2016mimic}\footnote{\url{https://physionet.org/content/mimiciii/1.4/}} covers 38,597 patients and 49,785 hospital admissions information in critical care units at the Beth Israel Deaconess Medical Center ranging from 2001 to 2012.
It includes deidentified administrative information such as demographics and highly granular clinical information, including vital signs, laboratory results, procedures, medications, caregiver notes, imaging reports, and mortality.

\subsection{eICU}
Similar to MIMIC-III, eICU~\citep{pollard_eicu_2018}\footnote{\url{https://physionet.org/content/eicu-crd/2.0/}} includes over 200,000 admissions from multiple critical care units across the United States in 2014 and 2015.
It contains deidentified administrative information following the US Health Insurance Portability and Accountability Act (HIPAA) standard and structured clinical data, including vital signs, laboratory measurements, medications, treatment plans, admission diagnoses, and medical histories.

\subsection{TREQS}
TREQS~\citep{wang2020text} is a healthcare question and answering benchmark that is built upon the MIMIC-III~\citep{johnson2016mimic} dataset. In TREQS, questions are generated automatically using pre-defined templates with the text-to-SQL task. Compared to the MIMIC-III dataset within the EHRSQL~\citep{lee2022ehrsql} benchmark, TREQS has a narrower focus in terms of the types of questions and the complexity of SQL queries. Specifically, it is restricted to only five tables but includes a significantly larger number of records (Table~\ref{tab:data}) within each table.

\section{Tool Set Details}
\label{app:toolset}
To obtain relevant information from EHRs and enhance the problem-solving capabilities of LLM-based agents, we augment LLMs with the following tools:

\noindent $\diamond$ \underline{\textbf{Database Loader}} loads a specific table from the database.

\noindent $\diamond$ \underline{\textbf{Data Filter}} applies specific filtering condition to the selected table. These conditions are defined by a column name and a relational operator. The relational operator may take the form of a comparison (e.g., "<" or ">") with a specific value, either with the column's values or the count of values grouped by another column.
Alternatively, it could be operations such as identifying the minimum or maximum values within the column.

\noindent $\diamond$ \underline{\textbf{Get Value}} retrieves either all the values within a specific column or performs basic operations on all the values, including calculations for the mean, maximum, minimum, sum, and count.

\noindent $\diamond$ \underline{\textbf{Calculator}} calculates the results from input strings. We leverage the WolframAlpha API portal\footnote{\url{https://products.wolframalpha.com/api}}, which can handle both straightforward calculations such as addition, subtraction, and multiplication and more complex operations like averaging and identifying maximum values.

\noindent $\diamond$ \underline{\textbf{Date Calculator}} calculates the target date based on the input date and the provided time interval information.

\noindent $\diamond$ \underline{\textbf{SQL Interpreter}} interprets and executes SQL code written by LLMs.


\section{Baseline Details}
\label{app:baseline_attr}
All the methods, including baselines and \method, share the same \emph{(1) tool definitions, (2) table meta information, and (3) few-shot demonstrations} in the prompts by default. The only difference is the prompting style or technical differences between different methods, which guarantees a \emph{fair comparison} among all baselines and \method.
Table~\ref{tab:baseline_attr} summarizes the inclusion of different components in both baselines and ours.

\begin{table*}[ht]
  {
  \resizebox{\linewidth}{!}{
  \begin{tabular}{lccccccc}
  \toprule
  \bfseries Baselines & \bfseries Tool Use & \bfseries \makecell{Code\\ Interface} & \bfseries \makecell{Environment\\ Feedback} & \bfseries Debugging & \bfseries \makecell{Error\\ Exploration} & \bfseries \makecell{Medical\\ Information} & \bfseries \makecell{Long-term\\ Memory} \\
  \midrule
  \multicolumn{7}{l}{\quad \emph{w/o Code Interface}}  \\ \midrule
  CoT~\citep{wei2022chain} & \color{green}{\ding{51}} & \color{red}{\ding{55}} & \color{red}{\ding{55}} & \color{red}{\ding{55}} & \color{red}{\ding{55}} & \color{red}{\ding{55}} & \color{red}{\ding{55}}\\
  Self-Consistency~\citep{wang2022self} & \color{green}{\ding{51}} & \color{red}{\ding{55}} & \color{red}{\ding{55}} & \color{red}{\ding{55}} & \color{red}{\ding{55}} & \color{red}{\ding{55}} & \color{red}{\ding{55}}\\
  Chameleon~\citep{lu2023chameleon} & \color{green}{\ding{51}} & \color{red}{\ding{55}} & \color{red}{\ding{55}} & \color{red}{\ding{55}} & \color{red}{\ding{55}} & \color{red}{\ding{55}} & \color{red}{\ding{55}}\\
  ReAct~\citep{yao2022react} & \color{green}{\ding{51}} & \color{red}{\ding{55}} & \color{green}{\ding{51}} & \color{red}{\ding{55}} & \color{red}{\ding{55}} & \color{red}{\ding{55}} & \color{red}{\ding{55}}\\
  Reflexion~\citep{shinn2023reflexion} & \color{green}{\ding{51}} & \color{red}{\ding{55}} & \color{green}{\ding{51}} & \color{green}{\ding{51}} & \color{red}{\ding{55}} & \color{red}{\ding{55}} & \color{red}{\ding{55}}\\\midrule
  \multicolumn{7}{l}{\quad \emph{w/ Code Interface}}  \\ \midrule
  LLM2SQL~\citep{nan2023evaluating} & \color{red}{\ding{55}} & \color{green}{\ding{51}} & \color{red}{\ding{55}} & \color{red}{\ding{55}} & \color{red}{\ding{55}} & \color{red}{\ding{55}} & \color{red}{\ding{55}}\\
  DIN-SQL~\citep{pourreza2024din} & \color{red}{\ding{55}} & \color{green}{\ding{51}} & \color{red}{\ding{55}} & \color{red}{\ding{55}} & \color{red}{\ding{55}} & \color{red}{\ding{55}} & \color{red}{\ding{55}}\\
  Self-Debugging~\citep{chen2023teaching} & \color{red}{\ding{55}} & \color{green}{\ding{51}} & \color{green}{\ding{51}} & \color{green}{\ding{51}} & \color{red}{\ding{55}} & \color{red}{\ding{55}} & \color{red}{\ding{55}}\\
  AutoGen~\citep{wu2023autogen} & \color{green}{\ding{51}} & \color{green}{\ding{51}} & \color{green}{\ding{51}} & \color{green}{\ding{51}} & \color{red}{\ding{55}} & \color{red}{\ding{55}} & \color{red}{\ding{55}}\\ 
  \rowcolor{gray!16} \textbf{\method (Ours) }& \color{green}{\ding{51}} & \color{green}{\ding{51}} & \color{green}{\ding{51}} & \color{green}{\ding{51}} & \color{green}{\ding{51}} & \color{green}{\ding{51}} & \color{green}{\ding{51}}\\
  \bottomrule
  \end{tabular}
  }}
\caption{Comparison of baselines and \method on the inclusion of different components.}
  \label{tab:baseline_attr}
\end{table*}

\noindent $\bullet$ \textbf{Baselines w/o Code Interface.} 
LLMs without a code interface rely purely on natural language-based planning capabilities. 

\noindent $\diamond$  \underline{\textbf{CoT}}~\citep{wei2022chain}: CoT enhances the complex reasoning capabilities of original LLMs by generating a series of intermediate reasoning steps.

\noindent $\diamond$  \underline{\textbf{Self-Consistency}}~\citep{wang2022self}: 
Self-consistency improves CoT by sampling diverse reasoning paths to replace the native greedy decoding and select the most consistent answer.

\noindent $\diamond$  \underline{\textbf{Chameleon}}~\citep{lu2023chameleon}: Chameleon employs LLMs as controllers and integrates a set of plug-and-play modules, enabling enhanced reasoning and problem-solving across diverse tasks.

\noindent $\diamond$  \underline{\textbf{ReAct}}~\citep{yao2022react}: ReAct integrates reasoning with tool use by guiding LLMs to generate intermediate verbal reasoning traces and tool commands.

\noindent $\diamond$  \underline{\textbf{Reflexion}}~\citep{shinn2023reflexion}:
Reflexion leverages verbal reinforcement to teach LLM-based agents to learn from linguistic feedback from past mistakes.

\noindent $\bullet$ \textbf{Baselines w/ Code Interface.}
LLMs with a code interface enhance the inherent capabilities of LLMs by enabling their interaction with programming languages and the execution of code.
In accordance with their default configuration, we present a summary of the utilization of programming languages in various baselines in Table~\ref{tab:coding}. Additionally, we provide a detailed explanation of the programming language selection in \method in Appendix~\ref{app:python}.

\noindent $\diamond$  \underline{\textbf{LLM2SQL}}~\citep{nan2023evaluating}: LLM2SQL augments LLMs with a code interface to generate SQL queries for retrieving information from EHRs for question answering.

\noindent $\diamond$
\underline{\textbf{DIN-SQL}}~\citep{pourreza2024din}: Compared to LLM2SQL, DIN-SQL further breaks down a complex problem into several sub-problems and feeding the solutions of those sub-problems into LLMs, effectively improving problem-solving performance. 

\noindent $\diamond$  \underline{\textbf{Self-Debugging}}~\citep{chen2023teaching}: 
Self-Debugging teaches LLMs to debug by investigating execution results and explaining the generated code in natural language.

\noindent $\diamond$  \underline{\textbf{AutoGen}}~\citep{wu2023autogen}: AutoGen unifies LLM-based agent workflows as multi-agent conversations and uses the code interface to encode interactions between agents and environments.
We follow the official tutorial\footnote{\url{https://microsoft.github.io/autogen/docs/Use-Cases/agent_chat/}} for the implementation of AutoGen. Specifically, we utilize the built-in \texttt{AssistantAgent} and \texttt{UserProxyAgent} within AutoGen to serve as the LLM agent and the code executor, respectively. The \texttt{AssistantAgent} is configured in accordance with AutoGen's tailored system prompts, which are designed to direct the LLM to (1) propose code within a coding block as required, (2) refine the proposed code based on execution outcomes, and (3) append a specific code, "TERMINATE", to conclude the response for terminating the dialogue. The \texttt{UserProxyAgent} functions as a surrogate for the user, extracting and executing code from the LLM's responses in a local environment. Subsequently, it relays the execution results back to the LLM. In instances where code is not detected, a standard message is dispatched instead. This arrangement facilitates an automated dialogue process, obviating the need for manual tasks such as code copying, pasting, and execution by the user, who only needs to initiate the conversation with an original query.

\begin{table}[ht]
\centering
\fontsize{8}{10}\selectfont\setlength{\tabcolsep}{0.5em}
\begin{tabular}{@{}lc@{}}
\toprule
Baselines       & \# Language     \\ \midrule
LLM2SQL~\citep{nan2023evaluating}     & SQL         \\
DIN-SQL~\citep{pourreza2024din}     & SQL                      \\ 
Self-Debugging~\citep{chen2023teaching} & SQL \\
AutoGen~\citep{wu2023autogen}        & Python       \\
\rowcolor{gray!16}\textbf{\method (Ours)} & \textbf{Python}  \\ \bottomrule
\end{tabular}
\caption{Comparison of baselines and \method on the selection of primary programming languages.}
\label{tab:coding}
\end{table}

\section{Selection of Primary Programming Language}
\label{app:python}
In our main experiments, we concentrate on three SQL-based EHR QA datasets to assess \method in comparison with other baselines. Nevertheless, we have opted for Python as the primary programming language for \method, rather than SQL\footnote{We include an empirical analysis in Appendix~\ref{app:codingexp} to further justify the selection of Python as the primary programming language for \method.}. The primary reasons for \emph{choosing Python instead of SQL} to address medical inquiries based on EHRs are outlined below:

\paragraph{Python Enables the External Tool-Use.}
Using alternative programming languages, such as SQL, can result in LLM-based agents becoming unavailable to external tools or functions.
The primary contribution of \method is to develop a code-empowered agent capable of generating and executing code-based plans to solve complex real-world clinic tasks. 
In general, the SQL language itself is incapable of calling API functions. For example, EHRXQA~\citep{bae2024ehrxqa} can be considered as an LLM agent that generates a solution plan in NeuralSQL (not SQL). This agent is equipped with two tools: a pre-trained Visual Question Answering (VQA) model called FUNC\_VQA, and a SQL interpreter. Similar to \method, it also relies on a non-SQL language and includes an SQL interpreter as a tool. Compared with NeuralSQL in EHRXQA~\citep{bae2024ehrxqa}, Python in \method can be directly executed, while NeuralSQL requires additional parsing.

\paragraph{Python Enables the Integration of SQL Tool Function.} 
Python provides excellent interoperability with various databases and data formats. It supports a wide range of database connectors, including popular relational databases such as PostgreSQL, MySQL, and SQLite, as well as non-relational databases like MongoDB. This interoperability ensures that \method can seamlessly interact with different EHR systems and databases.
Although our proposed method primarily relies on generating and executing Python code, \emph{we do not prohibit} \method \emph{from utilizing SQL to solve problems.} In our prompts and instructions, we also provide the 'SQLInterpreter' tool function for the agent to perform relational database operations using SQL. Through our experiments, we have observed that \method is capable of combining results from Python code and SQL commands effectively. For instance, when presented with the question, “Show me patient 28020's length of stay of the last hospital stay.”, \method will first generate SQL command \texttt{admit\_disch\_tuple = SQLInterpreter (\'SELECT ADMITTIME, DISCHTIME FROM admissions WHERE SUBJECT\_ID=28020 ORDER BY ADMITTIME DESC LIMIT 1\')} and execute it to obtain the tuples containing the patient's admission and discharge times. It will then employ Python code along with the built-in date-time function to calculate the duration of the last stay tuple.

\paragraph{Python Enables a More Generalizable Framework.} \method is a generalizable LLM agent empowered with a code interface to autonomously generate and execute code as solutions to given problems. While Section~\ref{sec:exp} focuses on the challenging multi-tabular reasoning task within EHRs for evaluation, the Python-based approach has the potential to be generalized to other tasks (e.g., risk prediction tasks based on EHRs) or even multi-modal clinical data and be integrated with additional tool-sets in the future. In contrast, other languages like SQL are limited to database-related operations.

\paragraph{Python is More Flexible in Extension.} Python is a general-purpose programming language that offers greater flexibility compared to SQL. It enables the implementation of complex logic and algorithms, which may be necessary for solving certain types of medical questions that require more than simple database queries. Python is also a highly flexible programming language that offers extensive capabilities through its libraries and frameworks, making it suitable for handling a wide range of programming tasks, including database operations. In contrast, SQL is only applicable within relational databases and does not provide the same level of flexibility and extension. This attribute is particularly important to LLM-based agents, as they can leverage both existing Python libraries and custom-defined functions as tools to solve complex problems that are inaccessible for and beyond the scope of SQL. 

\paragraph{Python Includes More Extensive Resources for Pre-Training.} 
Python has a large and active community of developers and researchers. This community contributes to the development of powerful libraries, frameworks, and tools that can be leveraged in \method. The extensive documentation, tutorials, and forums available for Python also provide valuable resources for troubleshooting and optimization.
\textbf{Github repositories are one of the most extensive sources of code data for state-of-the-art language models (\ie, LLMs), such as GPTs.} 
Python is the most widely used coding language on Github\footnote{\url{https://madnight.github.io/githut/\#/pull_requests/2023/1}}.
In addition, Python is known for its readability and maintainability. The clean and expressive syntax of Python makes it easier for researchers and developers to understand, modify, and extend the codebase of \method. This is particularly important when extended to real-world clinical research and practice, where the system may need to be updated frequently to incorporate new knowledge and adapt to evolving requirements.

\section{Additional Implementation Details}
\label{app:implementation}

\subsection{Hardware and Software Details}
All experiments are conducted on CPU: Intel(R) Core(TM) i7-5930K CPU @ 3.50GHz and GPU: NVIDIA GeForce RTX A5000 GPUs, using Python 3.9 and AutoGen 0.2.0\footnote{\url{https://github.com/microsoft/autogen}}. 

\subsection{Data Preprocessing Details}
During the data pre-processing stage, we create EHR question-answering pairs by considering text queries as questions and executing SQL commands in the database to automatically generate the corresponding ground-truth answers. We filter out samples containing unexecutable SQL commands or yielding empty results throughout this process.

\subsection{Code Generation Details}
Given that the majority of LLMs have been pre-trained on Python code snippets~\citep{gao2023pal}, and Python's inherent modularity aligns well with tool functions, we choose Python 3.9 as the primary coding language for interaction coding and AutoGen 0.2.0~\cite{wu2023autogen} as the interface for communication between the LLM agent and the code executor.

\subsection{Selection of Initial Set of Demonstrations}
The initial set of examples is collected manually, following four criteria: (1) using the same demonstrations across all the baselines; (2) utilizing all the designed tools; (3) covering as many distinct tables as possible; and (4) including examples in different styles of questions. With these criteria in mind, we manually crafted four demonstrations for each dataset. To ensure a fair comparison, we use the same initial four-shot demonstrations ($K = 4$) for all baselines and \method, considering the maximum length limitations of input context in baselines like ReAct~\citep{yao2022react} and Chameleon~\citep{lu2023chameleon}.

\subsection{Evaluation Metric Details}
Our main evaluation metric is the success rate (SR), quantifying the percentage of queries that the model successfully handles.
In addition, we leverage completion rate (CR) as a side evaluation metric to represent the percentage of queries for which the model is able to generate executable plans, regardless of whether the results are correct. 
Specifically, following existing LLM-based agent studies~\citep{xu2023tool,kirk2024improving}, we use CR to assess the effectiveness of LLM-based agents in generating complete executable plans without execution errors. One of our key components in \method is interactive coding with environmental feedback. By using CR, we can demonstrate that our proposed \method, along with other baselines that incorporate environmental feedback (e.g., ReAct~\citep{yao2022react}, Reflexion~\citep{shinn2023reflexion}, Self-Debugging~\citep{chen2023teaching}, and AutoGen~\citep{wu2023autogen}), has a stronger capability (higher CR) in generating complete executable plans without execution errors, compared to baselines without environmental feedback (e.g., CoT~\citep{wei2022chain}, Self-Consistency~\citep{wang2022self}, Chameleon~\citep{lu2023chameleon}, and LLM2SQL~\citep{nan2023evaluating}).


\subsection{EHR Metadata Details}
\noindent $\diamond$ \underline{\textbf{MIMIC-III.}}
\newline
\begin{Verbatim}
Read the following data descriptions, generate 
    the background knowledge as the context 
    information that could be helpful for 
    answering the question.
(1) Tables are linked by identifiers which 
    usually have the suffix 'ID'. For example, 
    SUBJECT_ID refers to a unique patient, 
    HADM_ID refers to a unique admission to 
    the hospital, and ICUSTAY_ID refers to a 
    unique admission to an intensive care unit.
(2) Charted events such as notes, laboratory 
    tests, and fluid balance are stored in a 
    series of 'events' tables. For example the 
    outputevents table contains all measurements
    related to output for a given patient, 
    while the labevents table contains 
    laboratory test results for a patient.
(3) Tables prefixed with 'd_' are dictionary 
    tables and provide definitions for 
    identifiers.c For example, every row of 
    chartevents is associated with a single 
    ITEMID which represents the concept 
    measured,  but it does not contain the 
    actual name of the measurement. By joining 
    chartevents and d_items on ITEMID, it is 
    possible to identify the concept represented 
    by a given ITEMID.
(4) For the databases, four of them are used to 
    define and track patient stays: admissions, 
    patients,icustays, and transfers. Another 
    four tables are dictionaries for cross-
    referencing codes against their respective 
    definitions: d_icd_diagnoses, 
    d_icd_procedures, d_items, and d_labitems. 
    The remaining tables, including chartevents, 
    cost, inputevents_cv, labevents, 
    microbiologyevents, outputevents, 
    prescriptions, procedures_icd, contain data 
    associated with patient care, such as 
    physiological measurements, caregiver 
    observations, and billing information.
\end{Verbatim}

\noindent $\diamond$ \underline{\textbf{eICU.}}
\newline
\begin{Verbatim}
Read the following data descriptions, generate 
    the background knowledge as the context 
    information that could be helpful for 
    answering the question.
(1) Data include vital signs, laboratory 
    measurements, medications, APACHE components, 
    care plan information, admission diagnosis, 
    patient history, time-stamped diagnoses 
    from a structured problem list, and 
    similarly chosen treatments.
(2) Data from each patient is collected into a 
    common warehouse only if certain interfaces 
    are available. Each interface is used to 
    transform and load a certain type of data: 
    vital sign interfaces incorporate vital 
    signs, laboratory interfaces provide 
    measurements on blood samples, and so on. 
(3) It is important to be aware that different 
    care units may have different interfaces in 
    place, and that the lack of an interface 
    will result in no data being available for 
    a given patient, even if those measurements 
    were made in reality. The data is provided 
    as a relational database, comprising 
    multiple tables joined by keys.
(4) All the databases are used to record 
    information associated to patient care, 
    such as allergy, cost, diagnosis, 
    intakeoutput, lab, medication, microlab, 
    patient, treatment, vitalperiodic.
\end{Verbatim}

\noindent $\diamond$ \underline{\textbf{TREQS.}}
\newline
\begin{Verbatim}
Read the following data descriptions, generate 
    the background knowledge as the context 
    information that could be helpful for 
    answering the question.
(1) The database contains five categories of 
    information for patients, including 
    demographics, laboratory tests, diagnosis, 
    procedures and prescriptions, and prepared 
    a specific table for each category 
    separately.
(2) These tables compose a relational patient 
    database where tables are linked through 
    patient ID and admission ID. 
\end{Verbatim}

\subsection{Prompt Details}
In the subsequent subsections, we detail the prompt templates employed in \method. The complete version of the prompts is available at our code repository due to space limitations.

\noindent $\diamond$ \underline{\textbf{Prompt for Code Generation.}}
We first present the prompt template for \method in code generation as follows:
\newline
\begin{Verbatim}
Assume you have knowledge of several tables:
{OVERALL_EHR_DESCRIPTIONS}
Write a python code to solve the given question. 
    You can use the following functions:
{TOOL_DEFINITIONS}
Use the variable 'answer' to store the answer 
    of the code. Here are some examples:
{4-SHOT_EXAMPLES}
(END OF EXAMPLES)
Knowledge:
{KNOWLEDGE}
Question: {QUESTION}
Solution: 
\end{Verbatim}

\noindent $\diamond$ \underline{\textbf{Prompt for Knowledge Integration.}}
We then present the prompt template for knowledge integration in \method as follows:
\newline
\begin{Verbatim}
Read the following data descriptions, generate 
    the background knowledge as the context 
    information that could be helpful for 
    answering the question.
{OVERALL_EHR_DESCRIPTIONS}
For different tables, they contain the 
    following information:
{COLUMNAR_DESCRIPTIONS}

{4-SHOT_EXAMPLES}

Question: {QUESTION}
Knowledge:
\end{Verbatim}

\noindent $\diamond$ \underline{\textbf{Prompt for `Rubber Duck' Debugging.}}
The prompt template used for debugging module in \method is shown as follows:
\newline
\begin{Verbatim}
Given a question:
{QUESTION}
The user has written code with the following 
    functions:
{TOOL_DEFINITIONS}

The code is as follows:
{CODE}

The execution result is:
{ERROR_INFO}

Please check the code and point out the most 
    possible reason to the error.
\end{Verbatim}

\noindent $\diamond$ \underline{\textbf{Prompt for Few-Shot Examples.}}
The prompt template used for few-shot examples in \method is shown as follows:
\newline
\begin{Verbatim}
Question: {QUESTION_I}
Knowledge:
{KNOWLEDGE_I}
Solution: {CODE_I}

Question: {QUESTION_II}
Knowledge:
{KNOWLEDGE_II}
Solution: {CODE_II}

Question: {QUESTION_III}
Knowledge:
{KNOWLEDGE_III}
Solution: {CODE_III}

Question: {QUESTION_IV}
Knowledge:
{KNOWLEDGE_IV}
Solution: {CODE_IV}
\end{Verbatim}

\section{Additional Experimental Results}
\subsection{Effect of Base LLMs}
\label{app:gpt-35}
Table~\ref{tab:gpt-35} presents a summary of the experimental results obtained from \method and all baselines using a different base LLM, \texttt{GPT-3.5-turbo (0613)}. The results clearly demonstrate that \method continues to outperform all the baselines, achieving a performance gain of 6.72\%.  This highlights the ability of \method to generalize across different base LLMs as backbone models.
When comparing the experiments conducted with \texttt{GPT-4} (Table~\ref{tab:main}), the performance of both the baselines and \method decreases. This can primarily be attributed to the weaker capabilities of instruction-following and reasoning in \texttt{GPT-3.5-turbo}.

\begin{table}[!ht]
\centering
\fontsize{7}{9}\selectfont\setlength{\tabcolsep}{0.2em}
\begin{tabular}{@{}l|cccc>{\columncolor{pink!10}}c>{\columncolor{blue!5}}c@{}}
\toprule
\textbf{Dataset ($\rightarrow$) }  & \multicolumn{6}{c}{\textbf{MIMIC-III}} \\ \midrule
\textbf{Complexity Level ($\rightarrow$)} & \textbf{I}     & \textbf{II}    & \textbf{III}   & \textbf{IV}    & \multicolumn{2}{c}{\textbf{All}}\\ 
\cmidrule{1-1}\cmidrule(lr){2-5} \cmidrule(lr){6-7}
\textbf{Methods ($\downarrow$) \slash Metrics ($\rightarrow$)} &  \multicolumn{4}{c}{\textbf{SR.}}  & \textbf{SR.} & \textbf{CR.} \\
\midrule
\multicolumn{7}{l}{\quad \emph{w/o Code Interface}}  \\\midrule 
CoT~\citep{wei2022chain}       &  23.16	&  10.40	&  2.99	&  1.71	&  8.62	&  41.55 \\
Self-Consistency~\citep{wang2022self} & 25.26	&  11.88	&  4.19	&  2.56	&  10.52	&  47.59 \\
Chameleon~\citep{lu2023chameleon}  & 27.37	& 11.88	& 3.59	& 2.56	& 11.21	& 47.59 \\
ReAct~\citep{yao2022react}      & 26.32 & 10.89	& 3.59	& 3.42	& 9.66	& 61.21     \\
Reflexion~\citep{shinn2023reflexion} & 30.53	& 12.38	& 9.58	& 8.55	& 13.28	& 66.72\\\midrule
\multicolumn{7}{l}{\quad \emph{w/ Code Interface}}  \\\midrule 
LLM2SQL~\citep{nan2023evaluating}    & 21.05	& 15.84	& 4.19	& 2.56	& 10.69	& 59.49         \\
Self-Debugging~\citep{chen2023teaching} & 36.84	& 33.66	& 22.75	& 16.24	& 27.59	& 72.93 \\
AutoGen~\citep{wu2023autogen}    & 28.42	& 25.74	& 13.17	& 10.26	& 19.48	& 52.42    \\
 \textbf{\method (Ours)}  & \textbf{43.16}	&\textbf{ 42.57}	& \textbf{29.94}	& \textbf{18.80}	& \textbf{34.31}	& \textbf{78.80} \\ \bottomrule
\end{tabular}
\caption{Experimental results of success rate (\ie, SR.) and completion rate (\ie, CR.) on MIMIC-III using \texttt{GPT-3.5-turbo} as the base LLM. The complexity of questions increases from Level I (the simplest) to Level IV (the most difficult). }\label{tab:gpt-35}
\end{table}

\subsection{Additional Ablation Studies}
\label{app:abl}
We conduct additional ablation studies to evaluate the effectiveness of each module in \method on eICU in Table~\ref{tab:ablation-eicu} and obtain consistent results.
From the results from both MIMIC-III and eICU, we observe that all four components contribute significantly to the performance gain. 

\noindent $\diamond$ \underline{\textbf{Medical Information Integration.}}
Out of all the components, the medical knowledge injection module mainly exhibits its benefits in challenging tasks. These tasks often involve more tables and require a deeper understanding of domain knowledge to associate items with their corresponding tables. 

\noindent $\diamond$ \underline{\textbf{Long-term Memory.}}
Following the reinforcement learning setting~\citep{sun2023adaplanner,shinn2023reflexion}, the long-term memory mechanism improves performance by justifying the necessity of selecting the most relevant demonstrations for planning. In order to simulate the scenario where the ground truth annotations (\ie, rewards) are unavailable, we further evaluate the effectiveness of the long-term memory on the completed cases in Table~\ref{tab:other-mem}, regardless of whether they are successful or not. The results indicate that the inclusion of long-term memory with completed cases increases the completion rate but tends to reduce the success rate across most difficulty levels, as some incorrect cases might be included as the few-shot demonstrations.  
We have also performed multi-round experiments with shuffled order and observed that the order had almost no influence on the final performance in all three datasets.
Nonetheless, it still outperforms the performance without long-term memory, confirming the effectiveness of the memory mechanism.

\noindent $\diamond$ \underline{\textbf{Interactive Coding.}}
For the ablation study setting of \method w/o interactive coding, we directly chose CoT~\citep{wei2022chain} as the backbone, where we deteriorate from generating code-based plans to \emph{natural language-based plans}. Once the steps are generated, we execute them in a step-by-step manner and \emph{obtain error information from the tool functions}. By combining the error messages with tool definitions and language-based plans, we are still able to prompt the LLMs to deduce the most probable underlying cause of the error. The medical information injection and long-term memory components remain unchanged from the original \method.
From the ablation studies, we can observe that
the interactive coding interface is the most significant contributor to the performance gain across all complexity levels. This verifies the importance of utilizing the code interface for planning instead of natural languages, which enables the model to avoid overly complex contexts and thus leads to a substantial increase in the completion rate. 
Additionally, the code interface also allows the debugging module to refine the planning with execution feedback, improving the efficacy of the planning process.

\noindent $\diamond$ \underline{\textbf{Debugging Module.}}
The `rubber duck' debugging module enhances the performance by guiding the LLM agent to figure out the underlying reasons for the error messages. This enables \method to address the intrinsic error that occurs in the original reasoning steps.
We then further illustrate the difference between debugging modules in \method and others.
Self-debugging~\citep{chen2023teaching} that sends back the execution results with an explanation of the code for plan refinement. Reflexion~\citep{shinn2023reflexion} sends the binary reward of whether it is successful or not back for refinement, which contains little information. In both cases, however, the error message is still information on the surface, like ‘incorrect query’, etc. This is aligned with our empirical observations that LLM agents tend to make slight modifications to the code snippets based on the error message without further debugging. Taking one step further, our debugging module in \method incorporates an error tracing procedure that enables the LLM to analyze potential causes beyond the current error message. Our debugging module aims to leverage the conversation format to think one step further about potential reasons, such as ‘incorrect column names in the query’ or ‘incorrect values in the query’. 

\begin{table}[ht]
\centering
\fontsize{8}{10}\selectfont\setlength{\tabcolsep}{0.3em}
\begin{tabular}{@{}lccc>{\columncolor{pink!10}}c>{\columncolor{blue!5}}c@{}}
\toprule
     Complexity level     & I     & II    & III     & \multicolumn{2}{c}{All}    \\
    \cmidrule(lr){2-4} \cmidrule(lr){5-6}
    Metrics &  \multicolumn{3}{c}{SR.}  & SR. & CR. \\
    \midrule
\textbf{\method}                         &     \textbf{54.82} & \textbf{53.52} & \textbf{25.00} & \textbf{53.10}      & \textbf{91.72} \\ 
\quad w/o medical information       & 36.75 & 28.39 & 6.25 &  30.17 & 47.24 \\
\quad w/o long-term memory                 & 52.41 & 44.22 & 18.75 & 45.69 & 78.97 \\ 
\quad w/o interactive coding & 46.39 & 44.97 & 6.25 & 44.31 & 65.34 \\
\quad w/o rubber duck debugging  & 50.60 & 46.98 & 12.50 & 47.07 & 70.86 \\
\bottomrule
\end{tabular}
\caption{Additional ablation studies on success rate (\ie, SR.) and completion rate (\ie, CR.) under different question complexity (I-III) on eICU dataset.}\label{tab:ablation-eicu}
\end{table}

\begin{table}[ht]
\centering
\fontsize{8}{10}\selectfont\setlength{\tabcolsep}{0.2em}
\begin{tabular}{@{}lcccc>{\columncolor{pink!10}}c>{\columncolor{blue!5}}c@{}}
\toprule
     Complexity level     & I     & II    & III   & IV    & \multicolumn{2}{c}{All}    \\
    \cmidrule(lr){2-5} \cmidrule(lr){6-7}
    Metrics &  \multicolumn{4}{c}{SR.}  & SR. & CR. \\
    \midrule
\textbf{\method} (LTM w/ Success)                              & 71.58 & \textbf{66.34} & \textbf{49.70}  & \textbf{49.14} & \textbf{58.97} & 85.86 \\ 
\quad LTM w/ Completion      & \textbf{76.84} & 60.89 & 41.92 & 34.48    & 53.24 & \textbf{90.05} \\
\quad w/o LTM                  & 65.96 & 54.46 & 37.13 & 42.74 & 51.73 & 83.42 \\ \bottomrule
\end{tabular}
\caption{Comparison on long-term memory (\ie, LTM) design under different question complexity (I-IV) on MIMIC-III dataset.}\label{tab:other-mem} 
\end{table}

\subsection{Cost Estimation}
\label{app:cost}
Using \texttt{GPT-4} as the foundational LLM model, we report the average cost of \method for each query in the MIMIC-III, eICU, and TREQS datasets as \$0.60, \$0.17, and \$0.52, respectively.
The cost is mainly determined by the complexity of the question (\ie, the number of tables required to answer the question) and the difficulty in locating relevant information within each table.

\section{Additional Empirical Analysis}




\subsection{Additional Question Complexity Analysis}
\label{app:qc-ana}
We further analyze the model performance by considering various measures of question complexity based on the number of elements in questions, and the number of columns involved in solutions, as shown in Figure~\ref{fig:qc}.
Incorporating more elements requires the model to either perform calculations or utilize domain knowledge to establish connections between elements and specific columns. 
Similarly, involving more columns also presents a challenge for the model in accurately locating and associating the relevant columns.
We notice that both \method and baselines generally exhibit lower performance on more challenging tasks\footnote{Exceptions may exist when considering questions of seven elements in Figures~\ref{fig:sr-qtag-m1} and~\ref{fig:cr-qtag-m1}, as it comprises only eight samples and may not be as representative.}. 
Notably, our model consistently outperforms all the baseline models across all levels of difficulty.
Specifically, for those questions with more than 10 columns, the completion rate of those open-loop baselines is very low (less than 20\%), whereas \method can still correctly answer around 50\% of queries, indicating the robustness of \method in handling complex queries with multiple elements.


\subsection{Additional Error Analysis}
\label{app:err}
We conducted a manual examination to analyze all incorrect cases generated by \method in MIMIC-III. Figure~\ref{fig:error} illustrates the percentage of each type of error frequently encountered during solution generation:

\noindent $\diamond$ \underline{\textbf{Date/Time.}} 
When addressing queries related to dates and times, it is important for the LLM agent to use the `Calendar' tool, which bases its calculations on the system time of the database. This approach is typically reliable, but there are situations where the agent defaults to calculating dates based on real-world time. Such instances may lead to potential inaccuracies.

\noindent $\diamond$ \underline{\textbf{Context Length.}} 
This type of error occurs when the input queries or dialog histories are excessively long, exceeding the context length limit.

\noindent $\diamond$ \underline{\textbf{Incorrect Logic.}} 
When solving multi-hop reasoning questions across multiple databases, the LLM agent may generate executable plans that contain logical errors in the intermediate reasoning steps. For instance, in computing the total cost of a hospital visit, the LLM agent might erroneously generate a plan that filters the database using \texttt{patient\_id} instead of the correct \texttt{admission\_id}.

\noindent $\diamond$ \underline{\textbf{Incorrect SQL Command.}} 
This error type arises when the LLM agent attempts to integrate the \texttt{SQLInterpreter} into a Python-based plan to derive intermediate results. Typically, incorrect SQL commands result in empty responses from \texttt{SQLInterpreter}, leading to the failure of subsequent parts of the plan.

\noindent $\diamond$ \underline{\textbf{Fail to Follow Instructions.}} 
The LLM agent often fails to follow the instructions provided in the initial prompt or during the interactive debugging process.

\noindent $\diamond$ \underline{\textbf{Fail to Debug.}} 
Despite undertaking all $T$-step trials, the LLM agent consistently fails to identify the root cause of errors, resulting in plans that are either incomplete or inexcusable.

\subsection{Additional Empirical Comparison of Primary Programming Languages}
\label{app:codingexp}
We conduct an additional analysis based on the empirical results (byond main results in Table~\ref{tab:main}) to further justify the selection of Python as our primary programming language.
\paragraph{Data Complexity.} 
The SPIDER~\citep{yu-etal-2018-spider} dataset, which is commonly used in SQL baselines~\citep{pourreza2024din}, typically only involves referencing information from \emph{an average of 1.1 tables per question}. In contrast, the EHRQA datasets we utilized require referencing information from \emph{an average of 1.9 tables per question}. This significant gap in \# tables\/questions indicates that EHRQA requires more advanced reasoning across multiple tables.
\paragraph{Sample Efficiency.}
SQL-based methods require more demonstrations. As SQL occupies a relatively smaller proportion of training data, it is quite difficult for LLMs to generate valid SQL commands. Usually, the methods need at least tens of demonstrations to get the LLMs familiar with the data schema and SQL grammar. In EHRAgents, we only need four demonstrations as few-shot multi-tabular reasoning.
\paragraph{Environment Feedback.} 
DIN-SQL~\citep{pourreza2024din} establishes a set of rules to automatically self-correct the SQL commands generated. Nevertheless, these rules are rigid and may not cover all potential scenarios. While it does contribute to enhancing the validity of the generated SQL commands to some extent, DIN-SQL lacks tailored information to optimize the code based on different circumstances, resulting in a lower success rate compared to self-debugging and \method, which provide error messages and deeper insights.
\paragraph{Execution Time Efficiency.}
We acknowledge that when handling large amounts of data, Python may experience efficiency issues compared to SQL commands. We have also observed similar challenges when working with the TREQS dataset, which contains a massive database with millions of records. However, in the MIMIC-III dataset, \method (avg. 52.63 seconds per question) still demonstrates higher efficiency compared to the state-of-the-art LLM4SQL method, DIN-SQL~\citep{pourreza2024din} (avg. 103.28 seconds per question). We will consider the efficiency of Python when dealing with large-scale databases as one of the important future directions.

\section{Additional Case Studies}
\label{app:case}

\begin{figure*}[ht]
  \centering
  \includegraphics[width=0.99\linewidth]{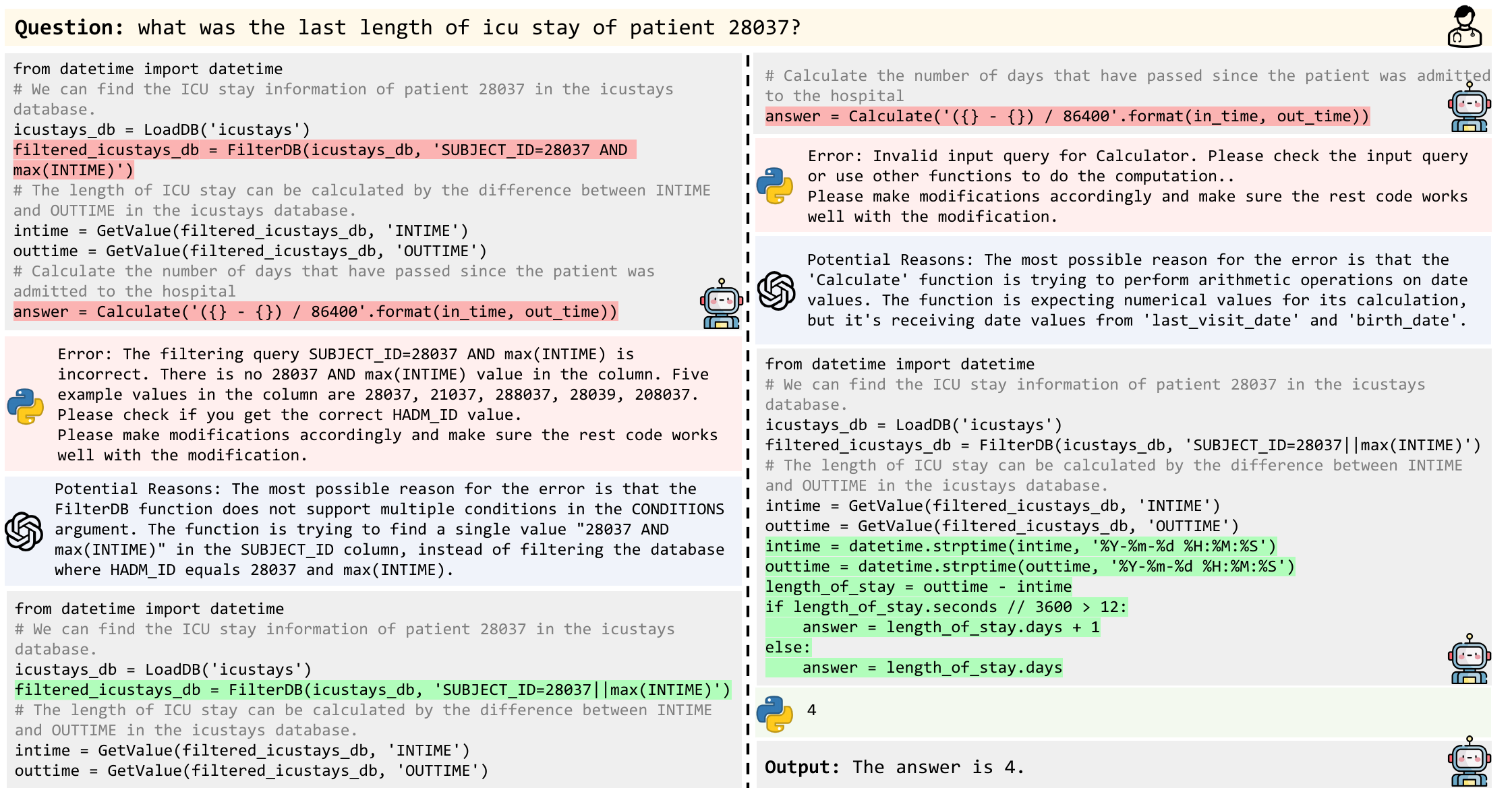}
  \vspace{-1ex}
  \caption{A complete version of case study in Figure~\ref{fig:cs-main1} showcasing interactive coding with environment feedback. 
  }
  \vspace{-1ex}
  \label{fig:debug-case3}
\end{figure*}

\begin{figure*}[ht]
  \centering
  \includegraphics[width=0.99\linewidth]{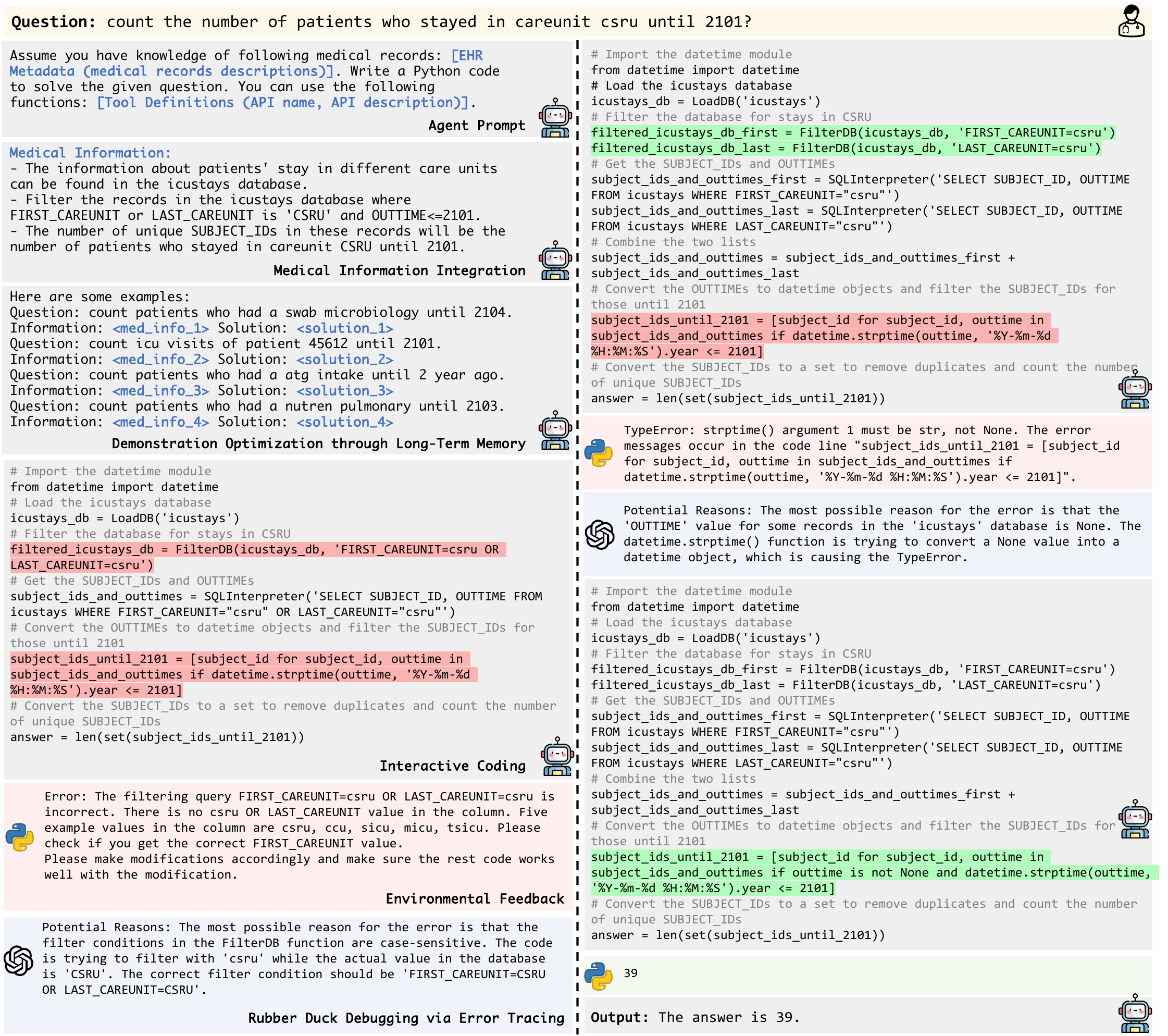}
  \vspace{-1ex}
  \caption{Case study of the complete workflow in \method. With EHR metadata and tool definitions, \method (1) integrates medical information to locate the required tables/records, (2) retrieves relevant examples from long-term memory, (3) generates and executes code, (4) iteratively debugs with error messages until the final solution.
  }
  \vspace{-1ex}
  \label{fig:debug-case2}
\end{figure*}

We present additional case studies to showcase the effectiveness of each module in \method, including medical information integration (Section~\ref{case:mii}), long-term memory (Section~\ref{case:memory}), code interface (Section~\ref{case:ci}), and rubber duck debugging module (Section~\ref{case:debug}). In addition, we include Figures~\ref{fig:debug-case3} and~\ref{fig:debug-case2} to showcase the entire workflow of \method with accumulative domain knowledge and coding capability for EHR reasoning.

\subsection{Case Study of Medical Information Integration}
\label{case:mii}

\begin{figure}[ht]
  \centering
  \includegraphics[width=0.99\linewidth]{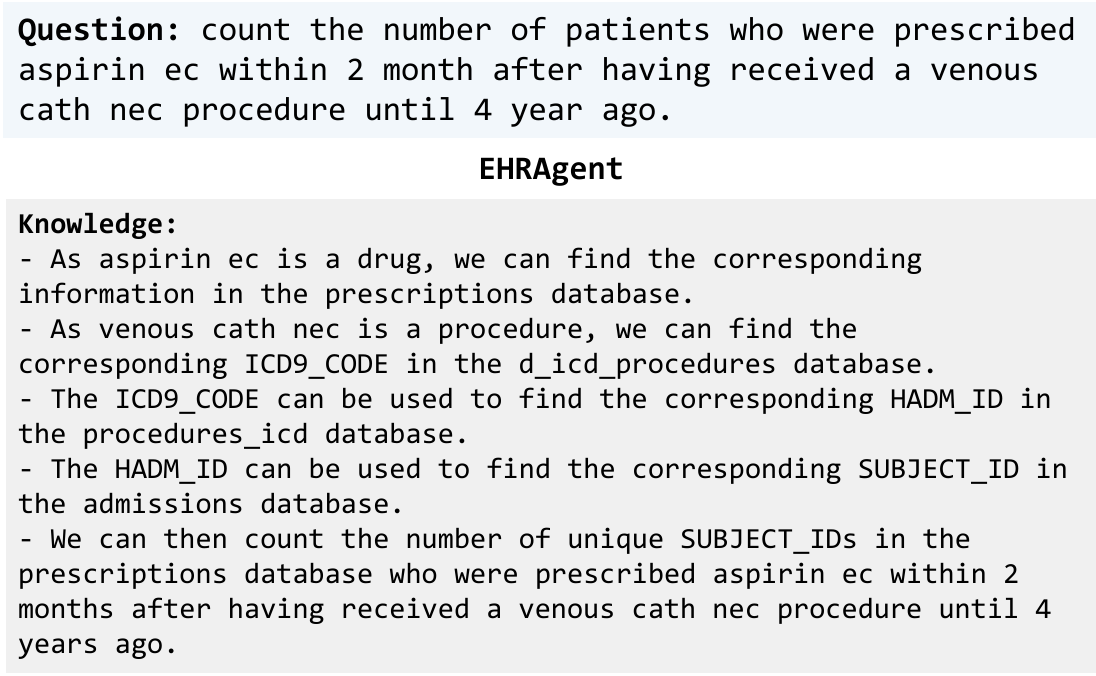}
  \caption{Case study of medical information injection in \method on MIMIC-III dataset. Given a question related to \textit{`aspirin ec'} and \textit{`venous cath nec'}, \method effectively integrates knowledge about their potential location in the database and the identifiers required to associate them.
  }
  \label{fig:cs-knowledge}
\end{figure}

Figure~\ref{fig:cs-knowledge} presents a case study of integrated medical knowledge based on a given query. When faced with a question related to \textit{`aspirin ec'} and \textit{`venous cath nec'}, \method effectively summarizes and integrates relevant knowledge containing drug and procedure information, successfully reasoning and identifying the potential locations of the required information (tables and records) in the EHR database (\eg, \textit{ `prescriptions'}, \textit{`d\_icd\_procedures'}), along with the detailed identifiers (\eg, \textit{`ICD9\_CODE'}, \textit{`HADM\_ID'}) needed to associate them. 

\subsection{Case Study of Long-Term Memory}
\label{case:memory}

\begin{figure*}[ht]
  \centering
  \includegraphics[width=0.99\linewidth]{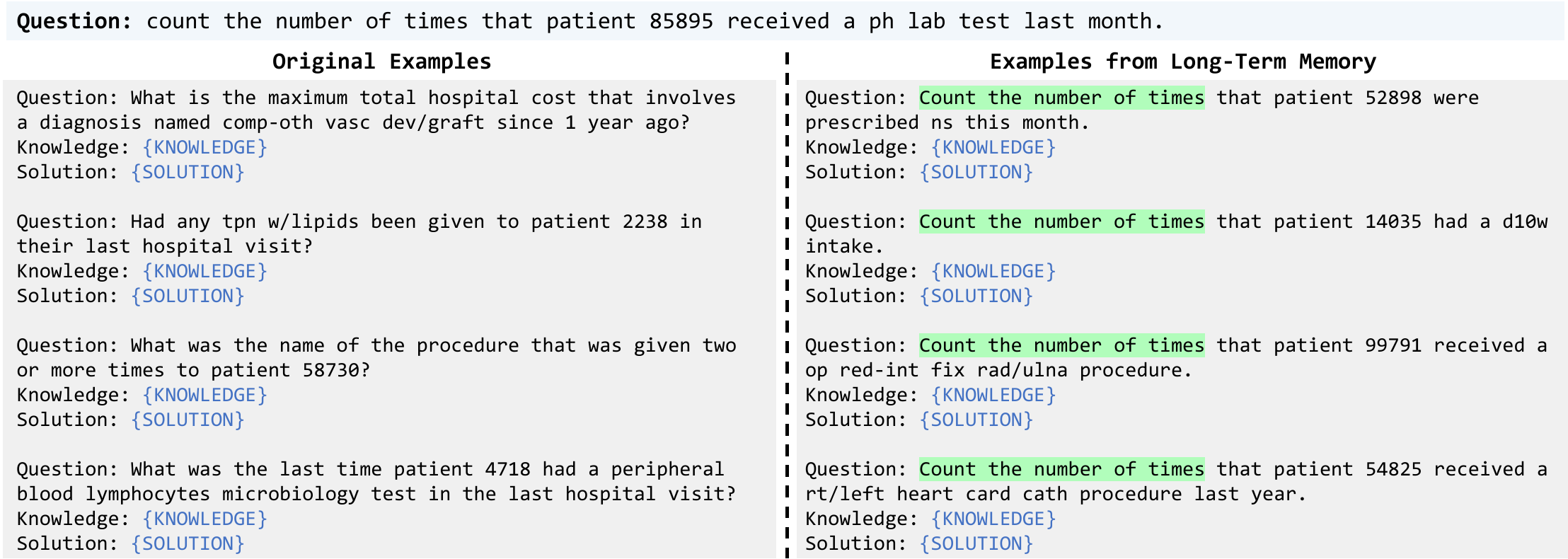}
  \caption{
  Case study of long-term memory in \method on MIMIC-III dataset.
  From the original few-shot examples on the left, none of the questions related to either `count the number' scenarios or procedure knowledge. In contrast, when we retrieve examples from the long-term memory, the new set is exclusively related to `count the number' questions, thus providing a similar solution logic for reference.
  }
  \label{fig:cs-memory}
\end{figure*}

Figure~\ref{fig:cs-memory} presents a case study of updating few-shot demonstrations from long-term memory.
Due to the constraints of limited context length, we are only able to provide a limited number of examples to guide \method in generating solution code. For a given question, the initial set of examples is pre-defined and fixed, which may not cover the specific reasoning logic or knowledge required to solve it.
Using long-term memory, \method replaces original few-shot demonstrations with the most relevant successful cases from past experiences for effective plan refinement.
For example, none of the original few-shot examples relate to either `count the number' scenarios or procedure knowledge; after selecting from the long-term memory pool, we successfully retrieve more relevant examples, thus providing a similar solution logic for reference.

\subsection{Case Study of Code Interface}
\label{case:ci}

\begin{figure*}[ht]
  \centering
  \includegraphics[width=0.99\linewidth]{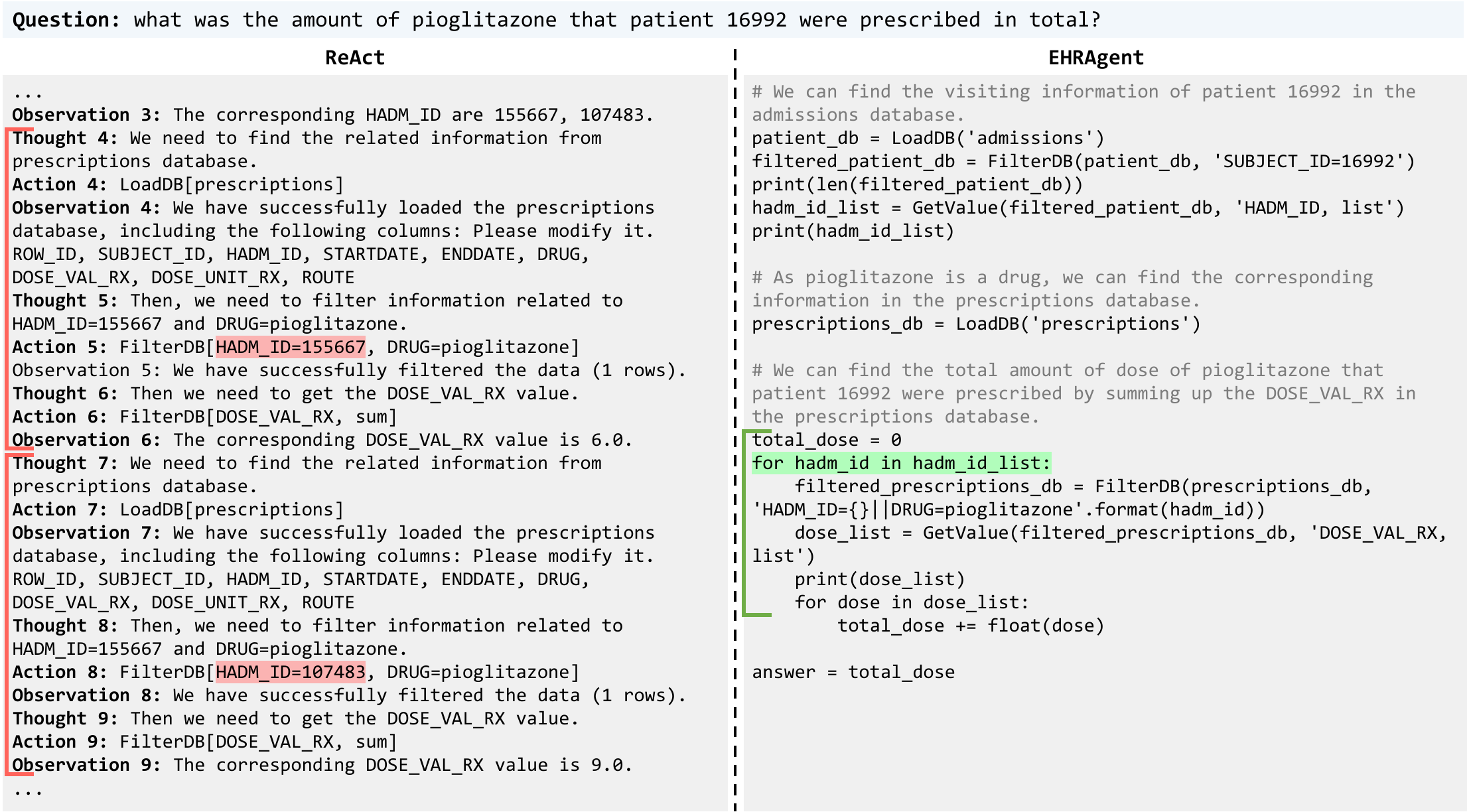}
  \caption{Case study 1 of code interface in \method on MIMIC-III dataset. The baseline approach, ReAct, lacks a code interface and encounters limitations when performing identical operations on multiple sets of data. It resorts to generating repetitive action steps iteratively, leading to an extended solution trajectory that may exceed the context limitations. In contrast, \method leverages the advantages of code structures, such as the use of `for loops', to address these challenges more efficiently and effectively. The steps marked in red on the left side indicate the repeated actions by ReAct, while the steps marked in green are the corresponding code snippets by \method. By comparing the length and number of steps, the code interface can help \method save much context space.
  }
  \label{fig:cs-code1}
\end{figure*}

\begin{figure*}[ht]
  \centering
  \includegraphics[width=0.99\linewidth]{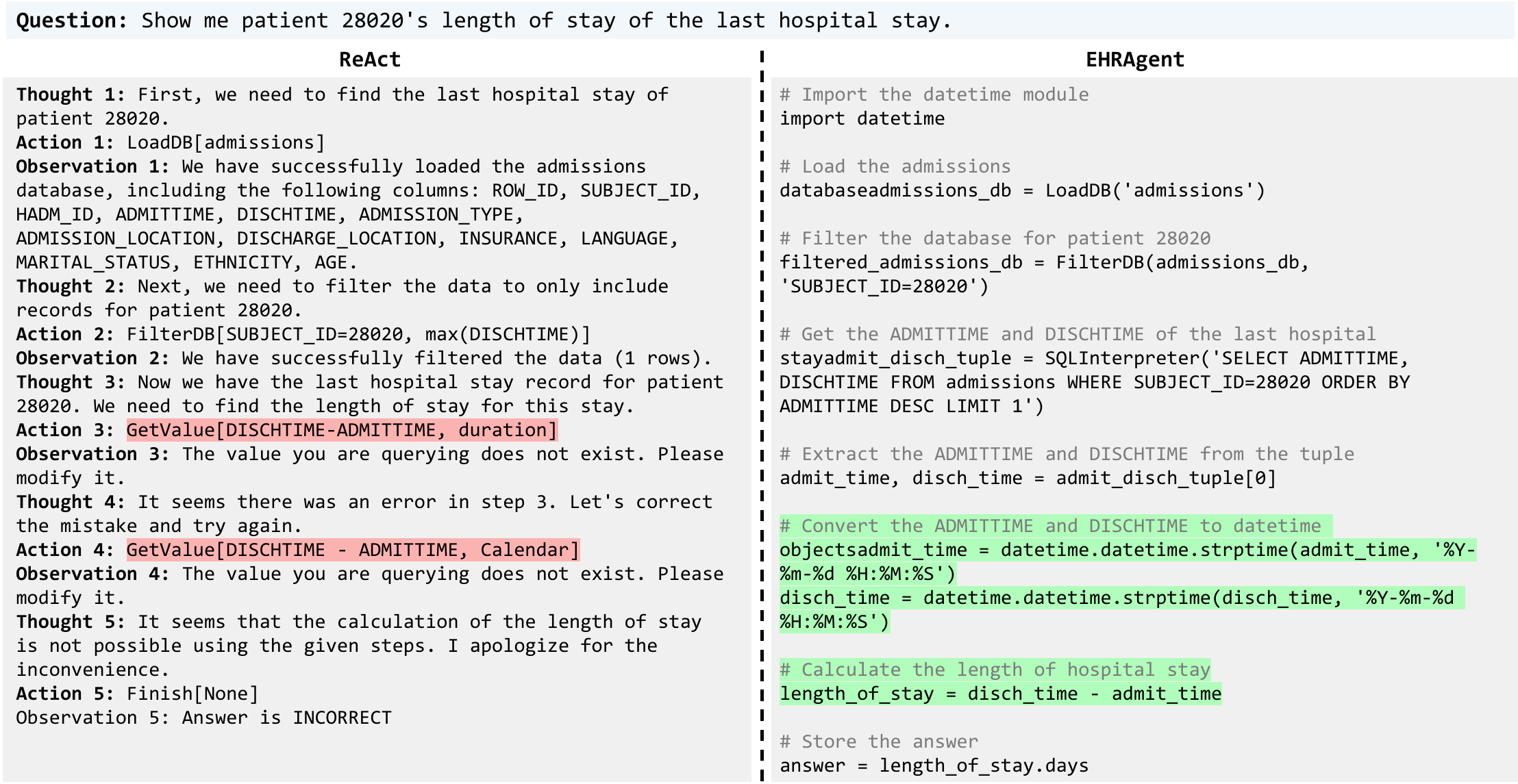}
  \caption{Case study 2 of code interface in \method on MIMIC-III dataset. When encountering challenges in tool use, ReAct will keep making trials and can be stuck in the modification process. On the other hand, with code interface, \method can take advantage of Python built-in functions to help with debugging and code modification.
  }
  \label{fig:cs-code2}
\end{figure*}

Figures~\ref{fig:cs-code1} and~\ref{fig:cs-code2} present two case studies of harnessing LLMs as autonomous agents in a multi-turn conversation for code generation, in comparison to a natural language-based plan such as ReAct. From the case studies, we can observe that ReAct lacks a code interface, which prevents it from utilizing code structures for efficient action planning and tool usage. This limitation often results in a lengthy context for ReAct to execute, which eventually leads to a low completion rate.

\subsection{Case Study of Rubber Duck Debugging}
\label{case:debug}
\begin{figure*}[ht]
  \centering
  \includegraphics[width=0.99\linewidth]{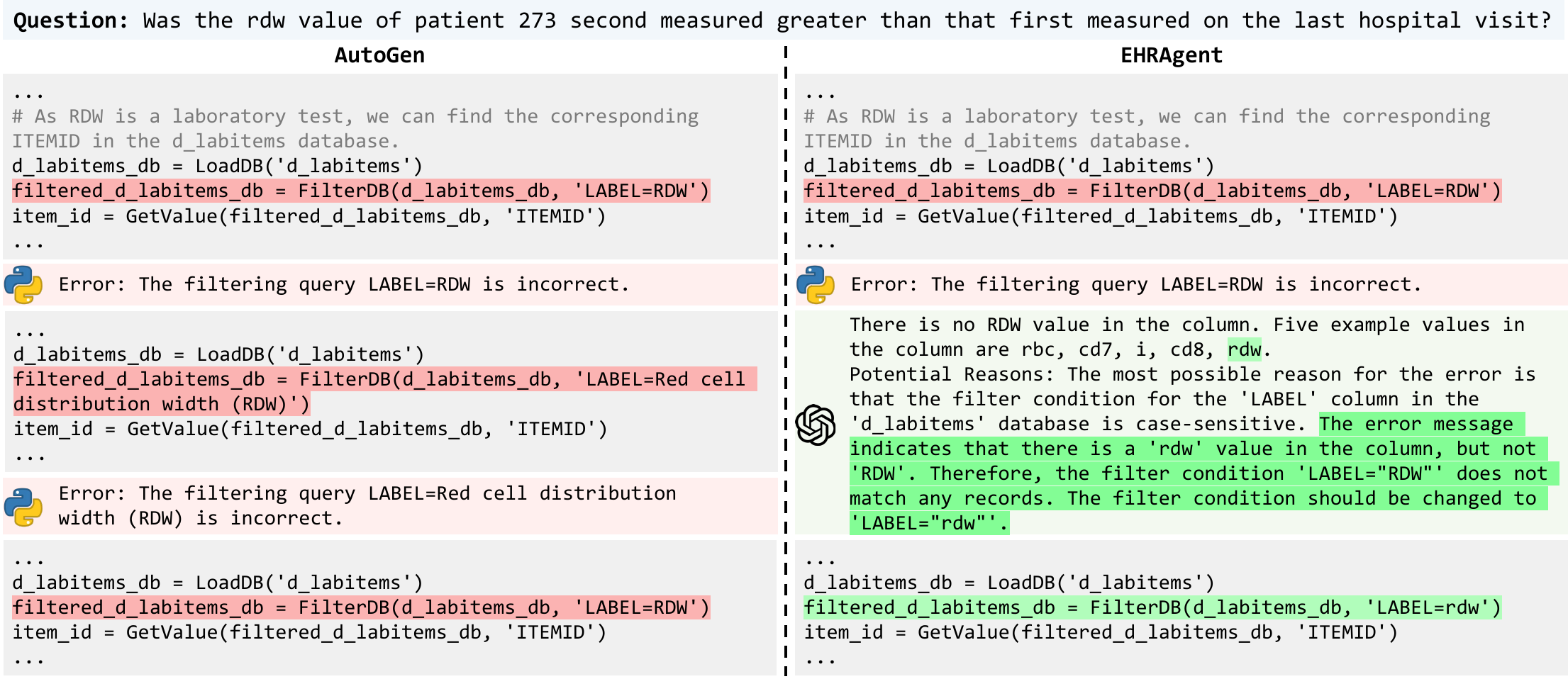}
  \caption{Comparative case study of the interactive coding process between AutoGen (\textit{left}) and \method (\textit{right}), where \method delves deeper into environmental feedback via debugging module to achieve plan refinement. 
  }
  \label{fig:cs-main}
\end{figure*}

Figure~\ref{fig:cs-main} showcases a case study comparing the interactive coding process between AutoGen and \method for the same given query. 
When executed with error feedback, AutoGen directly sends back the original error messages, making slight modifications (\eg, changing the surface string of the arguments) without reasoning the root cause of the error.
In contrast, \method can identify the underlying causes of the errors through interactive coding and debugging processes. It successfully discovers the underlying error causes (taking into account case sensitivity), facilitating accurate code refinement.

\end{document}